\documentclass[12pt]{article}

\usepackage[margin=1in]{geometry}
\usepackage{times}
\usepackage{setspace}
\usepackage{titlesec}
\titleformat{\section}{\bfseries\fontsize{12}{14}\selectfont}{\thesection.}{0.5em}{}
\titleformat{\subsection}{\bfseries\fontsize{11}{13}\selectfont}{\thesubsection}{0.5em}{}

\usepackage{amsmath, amssymb, amsfonts, amsthm}
\newtheorem{definition}{Definition}

\usepackage{graphicx}
\usepackage{caption}
\usepackage{subcaption}
\usepackage[table]{xcolor}
\usepackage{array}
\usepackage{tabularx}

\usepackage{natbib}
\setcitestyle{authoryear,open={(},close={)},aysep={}}
\usepackage{hyperref}
\hypersetup{
    colorlinks=true,
    urlcolor=blue,
    citecolor=black,
    linkcolor=black
}

\usepackage{fancyhdr}
\newcommand{\stitle}[1]{\noindent\textbf{#1}\quad}

\begin{document}


\vspace*{-0.25in}



\begin{center}
\colorbox{gray!20}{%
\parbox{\dimexpr\textwidth-2\fboxsep\relax}{%

\vspace{0.5em}

\centering
{\itshape\fontsize{10}{12}\selectfont
Proceedings of the International Conference on Industrial Engineering and Operations Management
\par}

\vspace{0.6em}

\begin{tabularx}{\textwidth}{@{}X>{\raggedleft\arraybackslash}X@{}}
{\fontsize{10}{12}\selectfont
Publisher: IEOM Society International, USA
}
&
{\fontsize{10}{12}\selectfont
DOI: \href{https://doi.org/10.46254/NA11.20260001}{10.46254/NA11.20260001}
}
\\[0.3em]
{\fontsize{10}{12}\selectfont
Published: June 9, 2026
}
&
{}
\end{tabularx}

\vspace{0.5em}

}%
}
\end{center}


\begin{center}
{\bfseries\fontsize{18}{20}\selectfont Building Trust in Black-box Optimization: A Comprehensive Framework for Explainability}
\end{center}

\vspace{1.25em}

\begin{center}
{\bfseries Nazanin Nezami} \\
University of Illinois Chicago \\
Chicago, Illinois, USA \\
nnezam2@uic.edu

\vspace{1em}

{\bfseries Hadis Anahideh} \\
University of Illinois Chicago \\
Chicago, Illinois, USA \\
hadis@uic.edu
\end{center}

\vspace{1.25em}

\begin{center}
{\bfseries Abstract}
\end{center}

\noindent
\fontsize{10}{12}\selectfont
Optimizing costly black-box functions within a constrained evaluation budget presents significant challenges in many real-world applications. Surrogate optimization (SO) is a common resolution; however, the complexity of surrogate models and the sampling core (e.g., acquisition functions) often introduces proprietary elements, leading to a lack of explainability and transparency. While existing literature has primarily focused on enhancing convergence to global optima, the practical interpretation of newly proposed strategies remains underexplored, particularly in batch evaluation settings. In this paper, we propose Inclusive Explainability Metrics for Surrogate Optimization (IEMSO), a comprehensive set of model-agnostic metrics designed to enhance the explainability of SO approaches. Through these metrics, we provide both intermediate and post-hoc explanations, enabling practitioners to build trust before and after conducting expensive evaluations. We consider four primary categories of metrics, each targeting a specific aspect of the SO process: Sampling Core Metrics, Batch Properties Metrics, Optimization Process Metrics, and Feature Importance Metrics. Our experimental evaluations demonstrate the significant potential of the proposed metrics across different benchmarks.

\vspace{1em}

\noindent
{\bfseries Keywords:} \\
\fontsize{10}{12}\selectfont
Black-box Optimization, Explainability Framework, Explainable Machine Learning

\vspace{1em}




\section{Introduction}\label{sec:intro}


Interpretability and reliability are foundational elements of any optimization approach \citep{carvalho2019machine}, serving as essential pillars upon which trust and acceptance are built. While explainability is a well-studied concept for machine learning (ML) models \citep{linardatos2020explainable}, the application of these principles to black-box optimization algorithms, such as Surrogate Optimization or Bayesian optimization \citep{rodemann2024explaining}, remains relatively unexplored. Black-box optimization (BBO) techniques aim to find the best parameter configurations or designs that optimize a target function, which is often unknown and costly to evaluate. The general idea is to sample the solution space strategically to discover promising areas.  

The importance of BBO cannot be overstated. It is widely used in various fields such as engineering design \citep{chen2019aerodynamic}, hyperparameter tuning for ML models \citep{turner2021bayesian}, and scientific lab experiments \citep{angermueller2020population}. These optimization techniques are crucial when the objective function is complex, expensive to evaluate, or lacks an explicit analytical form. With an efficient exploration and exploitation of the solution space, BBO helps identify optimal solutions that would be otherwise infeasible to find using traditional methods.


Despite their pivotal role in ML and other applications, black-box optimizers (acquisition functions) are frequently perceived as opaque and inscrutable. This lack of transparency hinders user trust and limits their broader applicability. Consequently, improving our understanding and interpretation of these optimization approaches can significantly enhance their usability and reliability in practical settings.

\emph{Example: The ``Robot Pushing'' problem involves two robot hands working together to push objects toward a designated goal. The movements of the robot's hands are influenced by 14 parameters, including hand trajectory, location, rotation, velocity, and direction. This complexity requires rigorous experimentation and precise optimization to achieve the desired outcomes. In this context, batch evaluation is crucial because it allows for efficient usage of computational resources and enables parallel processing. Evaluating a large batch of sample points simultaneously demands a transparent and explainable selection process.
Consider the scenario where a BBO algorithm selects a batch of 50 configurations for evaluation.
Engineers need to understand the rationale behind the choice of each configuration in the batch. This understanding is necessary to build confidence and trust in the optimizer's decisions and the overall optimization outcomes. For instance, engineers need to know if the selected configurations explore a diverse range of trajectories and velocities, or if they exploit the previously successful parameters. A transparent explanation of these choices can demonstrate how the algorithm balances exploration and exploitation, ensuring that the optimization process is effective and reliable.
}

Recent research has highlighted the importance of explainability in BBO. For instance, \citet{adachi2023looping} discussed that Bayesian optimization (BO) often struggles to gain user trust due to its opaque nature. They introduced the Collaborative and Explainable Bayesian Optimization (CoExBO) framework, which uses preference learning to incorporate human knowledge and Shapley values to quantify the contribution of each feature to the acquisition function.
Similarly, \citet{chakraborty2024explainable}
introduced TNTRules, a post-hoc, rule-based explainability method that employs multi-objective optimization to generate high-quality explanations for BO. TNTRules uses hierarchical agglomerative clustering (HAC) to uncover rules within the optimization space. It provides local explanations through visual graphs and actionable rules, and global explanations through ranked ``IF-THEN'' rules. These rules explain how the selected points relate to the optimization objectives and guide users in understanding the optimization process. \citet{rodemann2024explaining}
developed the ShapleyBO framework, which utilizes Shapley values to interpret proposed parameter configurations in BO. ShapleyBO quantifies each parameter's contribution to the configurations suggested by a BO algorithm based on their impact on the acquisition function. Shapley values help explain how each parameter influences this balance, providing transparency in the selection process. 

\stitle{Gaps and Contribution:} Despite recent advancements in enhancing the explainability of BBO, existing methods such as CoExBO, TNTRules, and ShapleyBO primarily focus on Bayesian optimization and do not generalize well to other surrogate optimization techniques, limiting their applicability. Additionally, they merely look at the contribution of features to the optimizer (often Gaussian Processes (GPs)), leaving a gap in comprehensive explainability for other types of surrogate models and optimization approaches. 
These limitations underscore the need for a more comprehensive approach to explainability in surrogate optimization. In response to these gaps, this paper proposes \emph{Inclusive Explainability Metrics for Surrogate Optimization (IEMSO)} to enhance the explainability of black-box optimization methods. Our proposed metrics are designed to be model-agnostic and can be applied to any framework, providing detailed explanations of various steps in surrogate optimization to improve human understanding and trust in the optimization process.

Our perspective and key contributions differ from existing explainable Bayesian optimization (BO) papers in several ways. Although a trivial application of our explainability metrics is for human collaboration with surrogate optimization, the primary focus is on gaining user trust rather than incorporating human feedback or learning human preferences, as emphasized in \citep{adachi2023looping}. Our metrics are applicable to any BO sampling strategy, not limited to specific acquisition function formats like those discussed in \cite{rodemann2024explaining}. Furthermore, instead of solely focusing on post-hoc explanations as in \citep{chakraborty2024explainable}, our proposed metrics offer both intermediate and post-hoc interpretations, aiding practitioners before and after performing expensive evaluations. Unlike \citep{chakraborty2024explainable}, which focuses on explaining the parameter space, we also address the explanation of single and batch BO proposals.

This paper introduces a wide range of explainability metrics applicable to different stages of surrogate optimization. Our proposed metrics are designed to be generalizable to any optimizers and surrogate models, thereby broadening the scope and utility of explainability in BBO. This comprehensive set of metrics ensures that the decision-making processes in surrogate optimization is explainable, enhancing their applicability across diverse optimization scenarios.
Our metrics are divided into four main categories, each focusing on a different aspect of the surrogate optimization process; 1) Sampling Core Metrics Explain the reasoning and strategy behind the sampling process.
2) Batch Properties Metrics Detail the criteria and efficiency of the generated sample batches.
3) Optimization Process Metrics Provide a comprehensive overview of the entire optimization process, including convergence and performance.
4) Feature Importance Metrics Describe the importance of features for the black-box objective, surrogate model behavior, and exploration or exploitation objectives.
This comprehensive set of metrics ensures that the decision-making processes in surrogate optimization is explainable, broadening the scope and utility of explainability in black-box optimization.

\section{Related Work}\label{sec:background}

Research in BBO has concentrated on managing the exploration-exploitation trade-off \citep{frazier2018tutorial}, traditionally in a sequential evaluation setting \citep{jones1998efficient,srinivas2009gaussian}. However, the introduction of batch sampling has created a paradigm shift, as it is not only necessary for numerous experimental evaluation scenarios but also accelerates convergence, enhances exploration, and improves robustness by leveraging parallel computational resources \citep{desautels2014parallelizing,balandat2020botorch}.
 
Explainable AI (XAI) is a growing field focused on making artificial intelligence systems more transparent and understandable to humans \citep{arrieta2020explainable}. Researchers have developed various techniques to elucidate the inner workings of complex models, ranging from local explanation methods such as LIME (Local Interpretable Model-agnostic Explanations) \citep{ribeiro2016should} and SHAP (SHapley Additive exPlanations) \citep{lundberg2017unified} to global interpretation strategies such as decision trees and rule-based systems \citep{linardatos2020explainable,sharma2024explainable}. However, research on explainability in black-box optimization is sparse. 
Only a few articles in the XAI literature have addressed the explainability of black-box optimizers. For example, \citet{seitz2022gradient} proposes a gradient-based explainability for GPs as the primary surrogate model in BO to create uncertainty-aware feature rankings. Explainable Metaheuristic \citep{singh2022towards} mines surrogate fitness models with the objective of identifying the variables that strongly influence solution quality for providing an explanation of the near-optimal solution. 
Recently, \citet{adachi2023looping} introduced Collaborative and Explainable Bayesian Optimization (CoExBo) specifically for lithium-ion battery design. This framework integrates human expertise into BO through preference learning and utilizes Shapley values to explain its recommendations. CoExBo begins by aligning human knowledge with BO via preference learning, followed by proposing multiple options from which the user can choose based on additional Shapley value insights. In contrast, Shapley-assisted human-BO collaboration in \citep{rodemann2024explaining} directly leverages Shapley values to align a single BO proposal with human input. Additionally, \citet{chakraborty2024explainable} recently presented TNTRules, a post-hoc rule-based explanation method for BO. This method identifies parameter subspaces for tuning using clustering algorithms, highlighting the advantages of XAI methods in collaborative BO settings.


\section{Inclusive Explainability Metrics for Surrogate Optimization}\label{sec:technical}

In this section, we introduce the Inclusive Explainability Metrics for Surrogate Optimization (IEMSO) framework, designed to enhance the explainability and trustworthiness of SO approaches. The IEMSO framework is structured around four primary categories of model-agnostic metrics, each targeting a specific aspect of the SO process. 


\subsection{Sampling Core Metrics}

This category of metrics focuses on providing explanations for individual sample points selected for expensive evaluation in each iteration. They offer insights into the logic behind the sampling process.
Sample points can be explained based on their contribution to the selected subset (batch), their location in the solution space, and their contribution to SO objective (exploration-exploitation trade-off).  

\noindent\stitle{Point Coordinate Exploration (PCE)} is a metric that quantifies the extent to which the selected sample points have covered the full range of each coordinate or feature within the predefined bounds.

\begin{definition}
Given a set of $n$ evaluated sample points $\mathcal{D}=\{\mathbf{x}^1, \mathbf{x}^2, \ldots, \mathbf{x}^n\}$, where each $\mathbf{x}^i$ is a $d$-dimensional vector and each coordinates $x_j$, $\forall j = 1, 2, \ldots, d$) is bounded by $[lb_j, ub_j]$, the PCE is defined as the normalized range covered by the sample points in that coordinate, ${PCE}_j = \frac{\max(x^1_{j}, \ldots, x^n_{j}) - \min(x^1_{j}, \ldots, x^n_{j})}{ub_j - lb_j}$. Hence, the average PCE is 
${PCE}(\mathcal{D}) = \frac{1}{d} \sum_{j=1}^{d} {PCE}_j$. 
\end{definition}


A ${PCE}$ value close to 1 indicates that the sample points have sufficiently explored the range of each coordinate $[lb_j, ub_j]$. Conversely, a ${PCE}$ value approaching 0 indicates that the sample points have explored only a limited fraction of each coordinate's range. This metric is simple to compute and offers a clear measure of coverage for each dimension. By normalizing exploration extent, PCE ensures consistency across different coordinates, helping to assess whether the sampling strategy is effectively exploring the search space which is a key factor in optimization.




\noindent\textbf{Mean Distance from Prior Evaluations (MDPE)} is a metric designed to quantify the dissimilarity (i.e., distance) of each individual sample point from the previously evaluated data points. In SO, sample points that exhibit greater distance from the previously evaluated points are predominantly important for the exploration objective, whereas points that are similar or in close proximity to the evaluated set are primarily selected for exploitation purposes.



\begin{definition}
Given a set of $n$ evaluated sample points $\mathcal{D}=\{\mathbf{x}^1, \mathbf{x}^2, \ldots, \mathbf{x}^n\}$, and the selected subset of candidate points $\mathcal{S}=\{\mathbf{z}^1, \mathbf{z}^2, \ldots, \mathbf{z}_k\}$, the MDPE is defined as the average pairwise Euclidean distance between each sample point $\mathbf{z}^i \in \mathcal{S}$ and the set of evaluated points $\mathcal{D}$, $MDPE(\mathbf{z}^k) = \frac{1}{n} \sum_{i=1}^{n} \|\mathbf{z}^k - \mathbf{x}^i\|$, where $\|\mathbf{z}^k - \mathbf{x}^i\|$ denotes the Euclidean distance between the points $\mathbf{z}^k$ and $\mathbf{x}^i$.
\end{definition}

The Mean Distance from Prior Evaluations (MDPE) metric offers several key advantages. Firstly, it is intuitive, providing a clear and understandable measure of how different or distant the new sample points are from the already evaluated points. This clarity aids in the easy interpretation of the sampling strategy's effectiveness. Secondly, MDPE effectively highlights the balance between exploration and exploitation, a crucial aspect of optimization tasks. Distant points indicate exploration, while points closer to previously evaluated ones indicate exploitation. Finally, MDPE offers a quantitative measure to assess the diversity and novelty of the selected sample points, enabling a rigorous evaluation of the optimization process.


\noindent\stitle{Contribution to the Hypervolume of Exploration-Exploitation (CHEE)} is a metric that quantifies the expected and actual impact of each sample point on exploration and exploitation objectives, both prior to and following the expensive evaluation.


\begin{definition}
Consider a surrogate optimization approach characterized by an exploitation metric $\mu(\mathbf{x})$ and an exploration metric $\sigma(\mathbf{x})$, where $\mathbf{x} \in \mathcal{X}$ represents a point in the solution space $\mathcal{X}$. Let $\hat{\mu}(\mathbf{x})$ and $\hat{\sigma}(\mathbf{x})$ denote the surrogate model's predictions for the exploitation and exploration metrics, respectively.
The estimated Pareto set $\mathcal{P}^{'}$ constructed on the estimated exploration and exploitation metrics is defined as $
\mathcal{P}^{'} = \left\{ \mathbf{x} \in \mathcal{X} \mid \nexists \mathbf{x}' \in \mathcal{X} {s.t.} \hat{\mu}(\mathbf{x}') \leq \hat{\mu}(\mathbf{x}) , \hat{\sigma}(\mathbf{x}') \leq \hat{\sigma}(\mathbf{x}) \right\}$.
The hypervolume (HV) of the Pareto set $\mathcal{P}^{'}$ with respect to a reference point $\mathbf{r}$ is $
HV(\mathcal{P}^{'}, \mathbf{r}) = \text{Volume} \left( \bigcup_{\mathbf{x} \in \mathcal{P}^{'}} [\mathbf{x}, \mathbf{r}] \right)$.
The contribution of a point $\mathbf{x} \in \mathcal{S}$ to the hypervolume of exploration-exploitation is then defined as $\nabla HV(\mathbf{x}, \mathcal{P}^{'}, \mathbf{r}) = HV(\mathcal{P}^{'}, \mathbf{r}) - HV(\mathcal{P}^{'} \setminus \{\mathbf{x}\}, \mathbf{r}).$
\end{definition}

This metric evaluates the change in hypervolume resulting from the inclusion or exclusion of a sample point, thus measuring its contribution to both exploration and exploitation. The use of hypervolume as a measure is a well-established method in multi-objective optimization to evaluate the quality of solutions. It provides a clear and interpretable way to assess the impact of adding or removing points from the Pareto set.
The exploration metric $\sigma(\mathbf{x})$ can be interpreted in different ways depending on the SO approach. When using a surrogate model, $\sigma(\mathbf{x})$ often represents the surrogate variance, which captures the model's uncertainty about the prediction at point $\mathbf{x}$. Alternatively, in a model-agnostic SO approach, $\sigma(\mathbf{x})$ can be represented by a distance metric $d(\mathbf{x})$, which measures the dissimilarity or novelty of the point $\mathbf{x}$ relative to other points in the dataset. This flexibility allows the framework to be adapted to various optimization scenarios and objectives.

\subsection{Batch Properties Metrics}
This category of metrics is dedicated to providing explanations for the selected batch of sample points. These metrics evaluate the batch based on its aggregated contribution to diversity (both intra-batch and inter-batch diversity), aggregated objective values, and its impact on the SO objective, specifically the exploration-exploitation trade-off.


\stitle{Density of the selected subset (DES)} is a metric that estimates the entropy of the batch. The entropy of a batch of sample points provides a measure of the uncertainty or randomness within the batch. In the context of surrogate optimization, higher entropy indicates greater diversity among the sample points, while lower entropy suggests that the points are more similar to each other. Entropy, in this context, can be calculated using techniques like Kernel Density Estimation (KDE)\citep{chen2017tutorial}, to estimate the underlying probability density function of a dataset without assuming a specific parametric form, making it suitable for complex distributions. 


\begin{definition}
For a batch of points $\mathcal{S} = \{\mathbf{x}_1, \mathbf{x}_2, \ldots, \mathbf{x}_k\}$, the DES metric can be defined using differential entropy. Let $\hat{p}(\mathbf{x})$ represent the KDE estimate of the probability density function for the points in $\mathcal{S}$. The differential entropy $H(\mathcal{S})$ is given by $
H(\mathcal{S}) = -\int_{\mathbb{R}^d} \hat{p}(\mathbf{x}) \log \hat{p}(\mathbf{x}) \, d\mathbf{x}$.  In practice, for a finite sample, the differential entropy can be approximated as $H(\mathcal{S}) \approx - \frac{1}{k} \sum_{i=1}^{k} \log \hat{p}(\mathbf{x}_i)$, where $\hat{p}(\mathbf{x}_i)$ is the KDE estimate at point $\mathbf{x}_i$.
\end{definition}

By calculating the differential entropy of the batch, the DES metric provides insights into how well the batch balances exploration and exploitation. High DES values indicate a well-distributed set of sample points, contributing to effective exploration of the solution space.

\noindent\stitle{Diversity of the selected subset (DIS)} quantifies the spread or coverage of the selected set of points in a multidimensional space by measuring the volume of the geometric shape spanned the points. Inspired by the diverse sampling concept from Determinantal Point Processes (DPPs) \citep{kulesza2012determinantal}, we aim to utilize the determinant of the kernel matrix $L$ to measure the batch diversity. 


\begin{definition}
Given a batch of points $\mathcal{S}$, the diversity can be measured using the determinant of the kernel matrix $L$, where $L_{ij}$ is computed using a similarity kernel function. $DIS(\mathcal{S})$ is then given by $DIS(\mathcal{S}) = \sum_{\sigma \in S_n} sign(\sigma) \prod_{i=1}^{n} L_{i, \sigma(i)}$, where $\sigma$ is a permutation of the indices ${1, 2, \ldots, n}$, and $\text{sign}(\sigma)$ is the sign of the permutation. When the kernel is a Radial Basis Function (RBF), $L_{ij} = \exp\left(-\frac{\|\mathbf{x}_i - \mathbf{x}_j\|^2}{2\sigma^2}\right)$, \citep{steinwart2006explicit}.
\end{definition}





\noindent\textbf{Average Batch Distance (ABD):} metric quantifies the dissimilarity between a batch of selected sample points and a set of previously evaluated data points. This metric is crucial for assessing the balance between exploration (distant points) and exploitation (close points).

\begin{definition}
Let $\mathcal{D} = \{\mathbf{x}_1, \mathbf{x}_2, \ldots, \mathbf{x}_n\}$ be a set of $n$ previously evaluated sample points, and let $\mathcal{S} = \{\mathbf{z}_1, \mathbf{z}_2, \ldots, \mathbf{z}_k\}$ be a batch of $k$ newly selected sample points. The ABD is defined as the average minimum Euclidean distance between each point in the selected batch $\mathcal{S}$ and the set of evaluated points $\mathcal{D}$: 
 $ABD(\mathcal{S}, \mathcal{D}) = \frac{1}{k} \sum_{i=1}^{k} \min_{j=1}^{n} \|\mathbf{z}_i - \mathbf{x}_j\|$,
where $\|\mathbf{z}_i - \mathbf{x}_j\|$ denotes the Euclidean distance between the points $\mathbf{z}_i$ and $\mathbf{x}_j$.

\end{definition}

\noindent\textbf{Hypervolume of Exploration-Exploitation (HVE)}: metric measures the volume of the region dominated by the sample points in the selected batch. Hypervolume is a widely used metric in multi-objective optimization \citep{emmerich2006single} to evaluate the quality of a set of points based on multiple objectives. We introduce this metric to explain the contribution of each batch to the exploration-exploitation trade-off in sampling for single-objective surrogate optimization.

\begin{definition}
Consider a surrogate optimization approach characterized by an exploitation metric $\mu(\mathbf{x})$ and an exploration metric $\sigma(\mathbf{x})$. Let $r = (r_\mu, r_\sigma^2)$ be the reference point in the 2D objective space, where $r_\mu$ represents the reference value for exploitation, set to a sufficiently high value worse than any expected $\mu(\mathbf{x})$ 
$r_\sigma^2$ represents the reference value for exploration, set to a sufficiently high value worse than any expected $\sigma^2(\mathbf{x})$. Let $\mathcal{S} = \{\mathbf{x}_1, \mathbf{x}_2, \ldots, \mathbf{x}_k\}$ be a batch of $k$ points selected in a given iteration. For each point $\mathbf{x}_i \in \mathcal{S}$, $\mu(\mathbf{x}_i)$ is the exploitation metric value at point $\mathbf{x}_i$ and $\sigma^2(\mathbf{x}_i)$ is the exploration metric value at point $\mathbf{x}_i$.
The hypervolume contribution $HV(\mathbf{x}_i)$ of each point $\mathbf{x}_i$ is defined as the volume of the rectangular region dominated by $\mathbf{x}_i$ and bounded by the reference point $r$, 
$HV(\mathbf{x}_i) = (r_\mu - \mu(\mathbf{x}_i)) \cdot (r_\sigma^2 - \sigma^2(\mathbf{x}_i))$. 
The total hypervolume contribution $HV_{\mathcal{S}}$ of the batch of $k$ points is the sum of the individual hypervolume contributions $HV_{\mathcal{S}} = \sum_{i=1}^{k} HV(\mathbf{x}_i) = \sum_{i=1}^{k} (r_\mu - \mu(\mathbf{x}_i)) \cdot (r_\sigma^2 - \sigma^2(\mathbf{x}_i))$.
\end{definition}

Similar to Definition 3, the exploration metric $\sigma(\mathbf{x})$ can be interpreted in different ways depending on the surrogate optimization (SO) approach. When using a surrogate model, $\sigma(\mathbf{x})$ often represents the surrogate variance, capturing the model's uncertainty about the prediction at point $\mathbf{x}$. Alternatively, in a model-agnostic SO approach, $\sigma(\mathbf{x})$ can be represented by a distance metric $d(\mathbf{x})$, measuring the dissimilarity or novelty of the point $\mathbf{x}$ relative to other points in the dataset.
This metric provides a measure of the combined contribution of the selected batch to both exploration and exploitation objectives, enabling the evaluation of the effectiveness of the sampling strategy in balancing these two aspects in surrogate optimization.

    


\subsection{Optimization Process Metrics}
These metrics provide a holistic view of the overall optimization process, including convergence and performance.

\stitle{Partitioning-Based Solution Space Analysis (PSSA)} metric focuses on providing an explanation of the solution space, which is commonly considered as a $d$-dimensional hyperrectangular space. Accurately locating each sample point or the selected batch of points within this space is crucial prior to expensive evaluations. To achieve this, we utilize partitioning methods to divide the solution space into distinct regions, facilitating a better understanding of the distribution and relationships within the data. 


\begin{definition}
Let $\mathcal{F} = \{(\mathbf{x}_i, y_i) \mid i = 1, 2, \ldots, n \}$ be a $d$-dimensional dataset, where $\mathbf{x}_i \in \mathbb{R}^d$ represents an evaluated sample point and $y_i$ is the corresponding expensive target value. A partitioning method divides the dataset into $m$ regions, $\{\mathcal{R}_1, \mathcal{R}_2, \ldots, \mathcal{R}_m\}$, where each region $\mathcal{R}_k$ is a subset of the solution space.
The region $\mathcal{R}_k$ can be defined as $
\mathcal{R}_k = \{\mathbf{x} \in \mathcal{X} \mid \text{Condition}_k(\mathbf{x})\}$,
where $\text{Condition}_k(\mathbf{x})$ specifies the criteria that determine the membership of $\mathbf{x}$ in region $\mathcal{R}_k$.
\end{definition}

As an example, consider decision trees as a partitioning method. In decision trees, the dataset is partitioned based on decision rules at each node, where each decision rule splits the data into two branches based on a threshold value of a specific feature. This process is repeated recursively, creating a hierarchical partitioning of the solution space.
Each node in the decision tree splits the data based on a decision rule of the form $\mathbf{x}_j \leq v$, where $\mathbf{x}_j$ is the $j$-th feature and $v$ is a threshold value. Each partition (leaf node) corresponds to a region $\mathcal{R}_k$ in the $d$-dimensional space. The region $\mathcal{R}_m$ associated with leaf node $m$ can be defined as $ \mathcal{R}_m = \left\{ \mathbf{x} \in \mathcal{X} \mid \bigwedge_{(j,v) \in \mathcal{L}_m} (\mathbf{x}_j \leq v \text{ or } \mathbf{x}_j > v) \right\},$ 
where $\mathcal{L}_m$ is the set of splitting rules leading to leaf $m$.

By leveraging the explainability of decision trees, the PSSA metric systematically partitions the solution space and derives interpretable rules that elucidate the relationships between the features and the target values. This approach enables a clearer understanding of the solution space and facilitates the identification of regions that warrant further exploration or exploitation.




\stitle{Convergence Rate (CR)} tracks the rate at which the optimization process approaches the optimal solution. It is measured by the decrease in the objective function value over iterations.

\begin{definition}
Given an objective function $f(\mathbf{x})$, let $\mathcal{I} = \{f(\mathbf{x}^{(1)}), f(\mathbf{x}^{(2)}), \ldots, f(\mathbf{x}^{(T)})\}$ be the sequence of best known objective function values over $T$ iterations. The Convergence Rate is defined as the average rate of decrease in the best known objective function value per iteration $CR = \frac{1}{T-1} \sum_{t=2}^{T} \left(\frac{f(\mathbf{x}^{(t-1)}) - f(\mathbf{x}^{(t)})}{f(\mathbf{x}^{(t-1)})}\right)$.

\end{definition}

This metric provides insight into how quickly the SO converges toward the optimal solution.

\stitle{Optimization Stability (OS)} metric evaluates the stability of the optimization process, ensuring it is not overly sensitive to initial conditions or random variations.

\begin{definition}
Let $\mathcal{S}$ be a set of $m$ optimization runs with different initial conditions or random seeds, resulting in objective function values $\{f(\mathbf{x}_1^*), f(\mathbf{x}_2^*), \ldots, f(\mathbf{x}_n^*)\}$ at convergence. The Optimization Stability is defined as the standard deviation of the final objective function values at convergence $OS = \sqrt{\frac{1}{m} \sum_{i=1}^{m} \left(f(\mathbf{x}_i^*) - \overline{f}\right)^2}$, where $\overline{f}$ is the mean of the objective function values at convergence $ \overline{f} = \frac{1}{n} \sum_{i=1}^{n} f(\mathbf{x}_i^*)$.

\end{definition}

This metric quantifies the variability in the optimization outcomes, indicating the robustness.



\subsection{Feature Importance}
This group of metrics focuses on evaluating the significance of individual features in the context of surrogate optimization. By quantifying the contribution of each feature to various objectives and models, these metrics provide insights into the behavior and performance of the optimization process, enhancing its interpretability.

\stitle{Feature Importance for Exploration and Exploitation (FIEE)} is a metric that evaluates the contribution of each feature to the exploration and exploitation objectives.

\begin{definition}
Given a set of input features $\mathbf{x} = (x_1, x_2, \ldots, x_d)$, the contribution of each feature $x_j$, $\forall j \in \{1, \ldots, d\}$, to the exploration objective is denoted by $\vert \eta_j \vert$, and to the exploitation objective by $\vert \lambda_j \vert$. These contributions can be calculated using feature importance techniques such as SHAP values, permutation importance, or other model-specific methods.
$$
\eta_j = \text{SHAP}_{\sigma}(x_j) \quad \text{or} \quad \eta_j = \text{Permutation Importance}_{\sigma}(x_j),
$$
$$
\lambda_j = \text{SHAP}_{\mu}(x_j) \quad \text{or} \quad \lambda_j = \text{Permutation Importance}_{\mu}(x_j),
$$

where $\text{SHAP}_{\sigma}(x_j)$ and $\text{Permutation Importance}_{\sigma}(x_j)$ represent the SHAP value and permutation importance of feature $x_j$ with respect to the exploration metric $\sigma(\mathbf{x})$, respectively, and $\text{SHAP}_{\mu}(x_j)$ and $\text{Permutation Importance}_{\mu}(x_j)$ represent the SHAP value and permutation importance of feature $x_j$ with respect to the exploitation metric $\mu(\mathbf{x})$, respectively.
\end{definition}



\stitle{Feature Importance for the Black-Box Objective Function (FIBB)} is a metric that evaluates the contribution of each feature to the black-box objective function.

\begin{definition}
The contribution of each feature $j \in \{1, \ldots, d\}$ to the black-box objective function is denoted by $\vert \phi_j \vert$. This contribution can be calculated using data-driven feature importance methods such as SHAP values, permutation importance, or other relevant techniques, 
$$
\phi_j = \text{SHAP}_{f}(x_j) \quad \text{or} \quad \phi_j = \text{Permutation Importance}_{f}(x_j),
$$

where $\text{SHAP}_{f}(x_j)$ and $\text{Permutation Importance}_{f}(x_j)$ represent the SHAP value and permutation importance of feature $x_j$ with respect to the black-box objective function $f(\mathbf{x})$.
\end{definition}

These metrics help in understanding the relative importance of features and their evolution across iterations, enhancing interpretability.


\stitle{Feature Importance of Surrogate Model (FIS)} is a metric that quantifies the contribution of individual features to the predictions generated by the surrogate model. It utilizes point-wise feature importance techniques, such as Shapley Values \citep{lundberg2017unified}, to offer a detailed analysis of how each feature influences the predicted outcomes.

\begin{definition}
Given a surrogate model $\hat{f}$ and a set of input features $\mathbf{x} = (x_1, x_2, \ldots, x_d)$, the contribution of each feature $j$ to the surrogate model’s prediction for a data point $\mathbf{x}_i$ can be quantified using feature importance scores $\phi_j(\mathbf{x}_i)$. The \textbf{FIS} metric for each feature $j$ is then defined as the average feature importance score over all data points $FIS_j = \frac{1}{n} \sum_{i=1}^{n} \phi_j(\mathbf{x}_i)$, where $\phi_j(\mathbf{x}_i)$ is the feature importance score of feature $j$ for the $i$-th data point, and $n$ is the total number of data points. Feature importance scores $\phi_j(\mathbf{x}_i)$ can be obtained using various techniques, such as SHAP values, permutation importance, or other model-specific importance measures like Multivariate Adaptive Regression Splines (MARS) \citep{friedman1991multivariate} and Decision Trees (DT) \citep{song2015decision}.
\end{definition}


We emphasize the advantages of using explainable surrogate models such as MARS or DT compared to other models like Gaussian Processes (GPs) \citep{rasmussen2010gaussian}. Although GPs are powerful for capturing complex, non-linear relationships, they lack inherent mechanisms for directly extracting feature importance. 

\section{Results}\label{sec:result}

In this section, we elaborate on the impact and applicability of the proposed explanation metrics, employing the well-known synthetic and real-world benchmark problems from black-box optimization literature.  

\noindent\textbf{Baselines:}
We consider the standard Monte-Carlo-based batch Bayesian acquisition functions such as \textbf{qEI}, \textbf{qUCB}, \textbf{qMES}, and \textbf{qGibbon} as described in \citep{wilson2017reparameterization,balandat2020botorch}. Moreover, we consider \textbf{DEEPA} \citep{nezami2024dynamic} as a Pareto-based SO baseline which has been shown to be effective for high-dimensional BBO problems. 

\noindent\textbf{Experiment Set-up:}
In our experiments, the size of the initial evaluated set is assigned based on the $2*(d+1)$ formula, where $d$ refers to the number of features. 
For the $10d$ test problems, we maintain a batch size of $k=8$, a candidate set of $1000$ randomly generated sample points.  
For the $2d$ instances, we use a batch size of $k=4$, a candidate set of $100$ randomly generated sample points. 
For the Levy $6d$ test problem, the batch size is set to $4$, and the candidate set includes $1000$ points. 
For the Robot Pushing and Rover Trajectory benchmarks, the batch size is set to $k=50$.

\begin{figure*}[!]
\centering

\begin{subfigure}{0.8\textwidth}
    \centering
    \includegraphics[width=\linewidth]{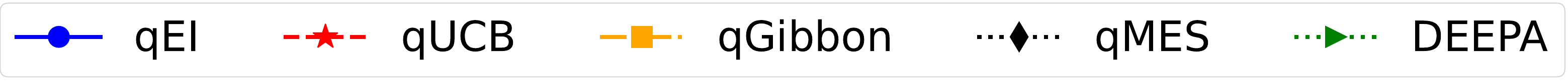}
\end{subfigure}

\vspace{0.5em}

\begin{subfigure}{0.4\textwidth}
    \centering
    \includegraphics[width=\linewidth]{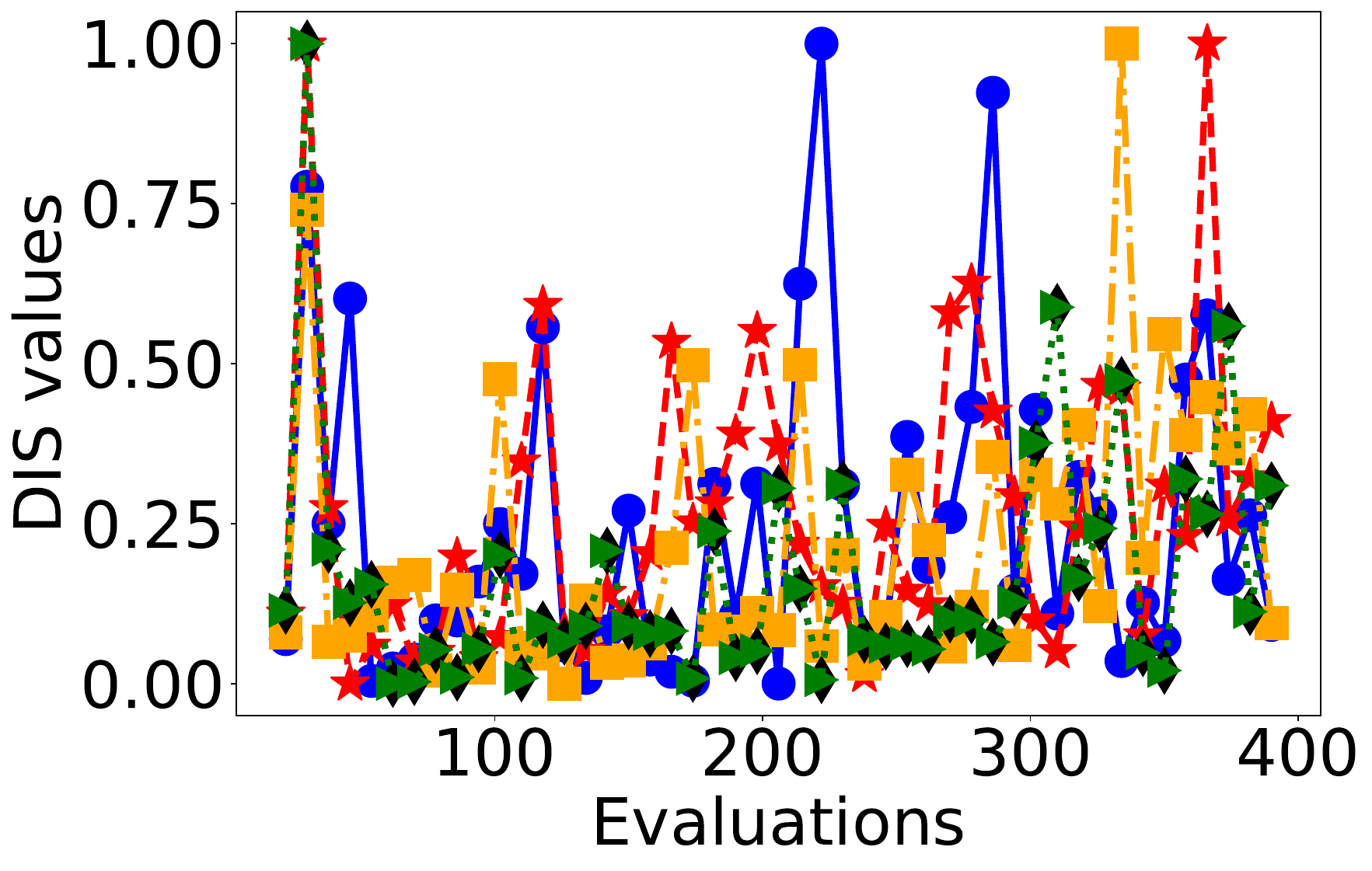}
    \caption{DIS}
    \label{fig:batch_rast}
\end{subfigure}
\begin{subfigure}{0.4\textwidth}
    \centering
    \includegraphics[width=\linewidth]{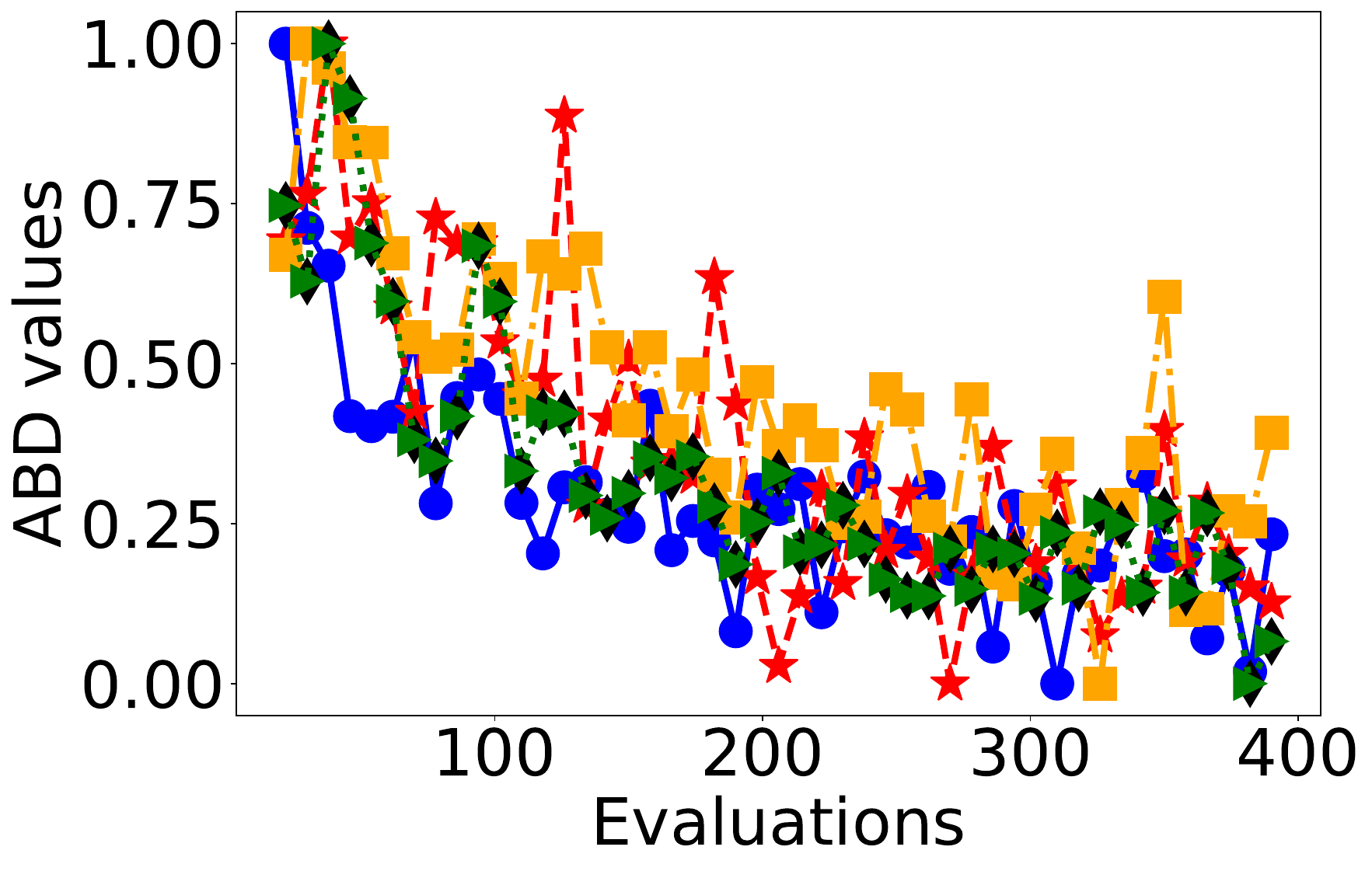}
    \caption{ABD}
    \label{fig:batch_robot}
\end{subfigure}

\caption{Batch Properties—Rastrigin 10D}
\label{fig:batchproperties1}
\end{figure*}

\begin{figure*}[!]
\centering

\begin{subfigure}{0.8\textwidth}
    \centering
    \includegraphics[width=\linewidth]{plots/Rast_leg.pdf}
\end{subfigure}

\vspace{0.5em}

\begin{subfigure}{0.4\textwidth}
    \centering
    \includegraphics[width=\linewidth]{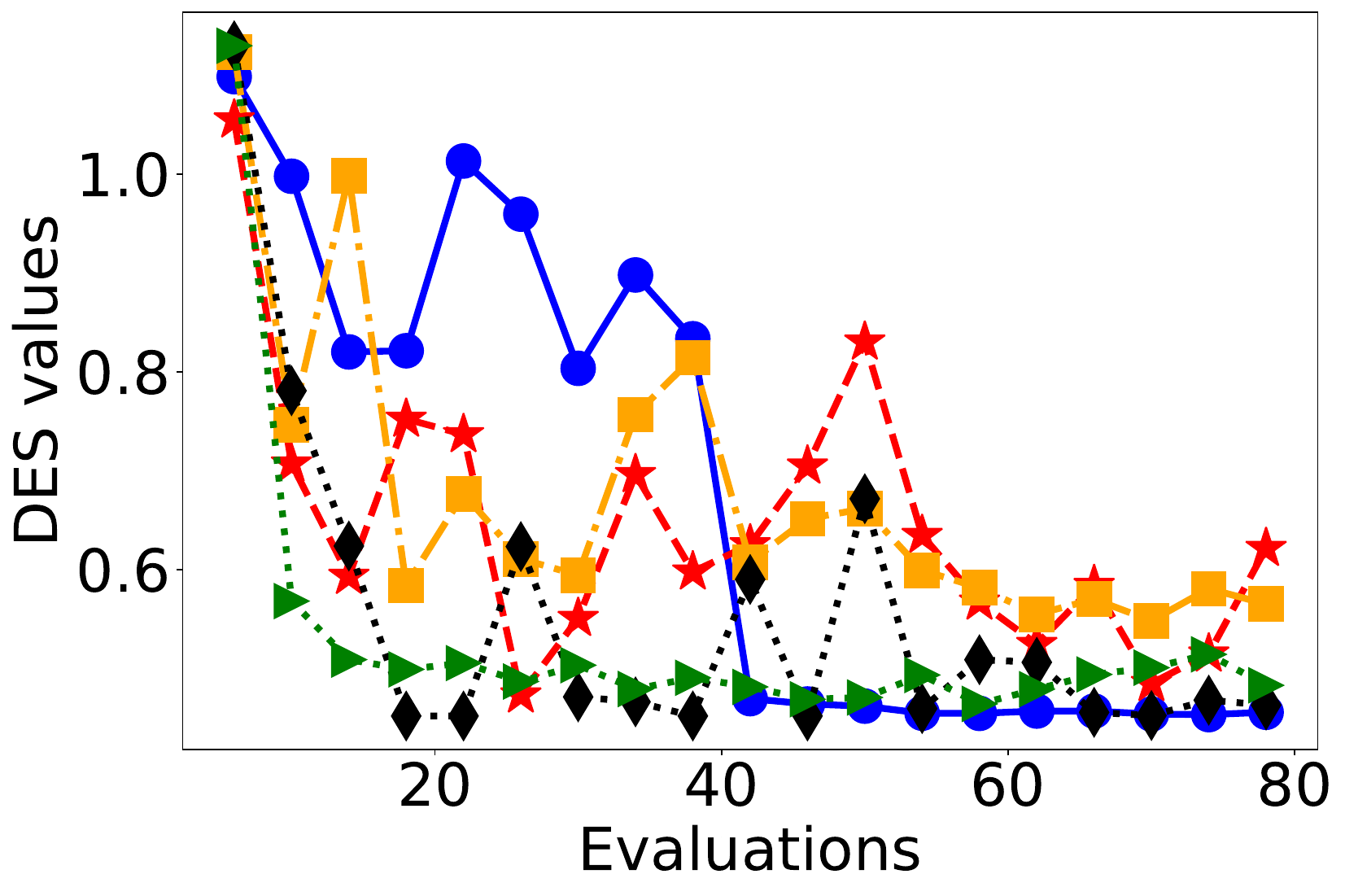}
    \caption{DES}
    \label{fig:CR}
\end{subfigure}
\begin{subfigure}{0.4\textwidth}
    \centering
    \includegraphics[width=\linewidth]{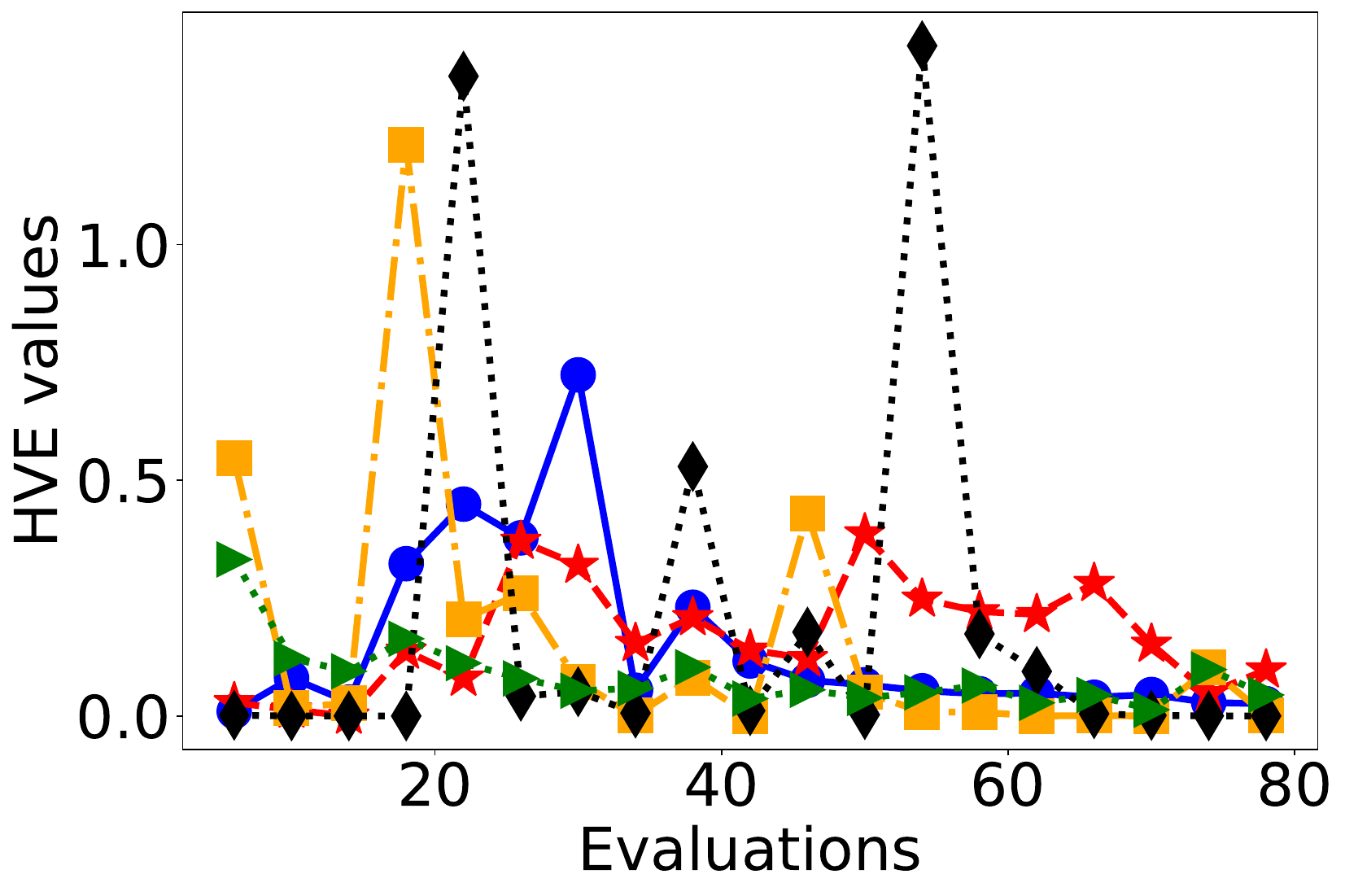}
    \caption{HVE}
    \label{fig:OS}
\end{subfigure}

\caption{Batch Properties—Rosenbrock 2D}
\label{fig:batchproperties2}
\end{figure*}

Figures~\ref{fig:batchproperties1} and ~\ref{fig:batchproperties2} demonstrate how \textbf{Batch Properties Metrics} provide valuable insights, enabling the user to compare batch selections across different baselines.
For the 10-dimensional Rastrigin test problem, the ABD plot indicates that \textbf{qGibbon} tends to select points that are more distant from the evaluated set for batch evaluation compared to other baselines. This characteristic makes \textbf{qGibbon} an excellent choice for users prioritizing exploration. This behavior aligns with the Gibbon acquisition function, which is equivalent to a Determinantal Point Process (DPP) when specific quality functions and similarity kernels are used. Maximizing the determinant of the selected subset ensures that a diverse set of points is chosen for evaluation.

\begin{figure}[!]
\centering

\begin{subfigure}{0.45\textwidth}
    \centering
    \includegraphics[width=\linewidth]{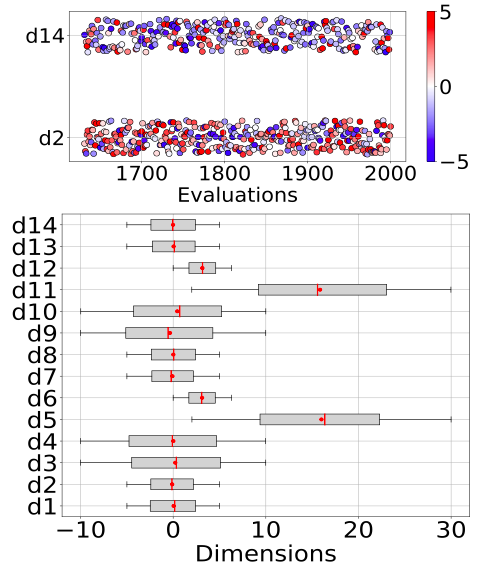}
    \caption{PCE}
    \label{fig:PCE_robot}
\end{subfigure}
\begin{subfigure}{0.45\textwidth}
    \centering
    \includegraphics[width=\linewidth]{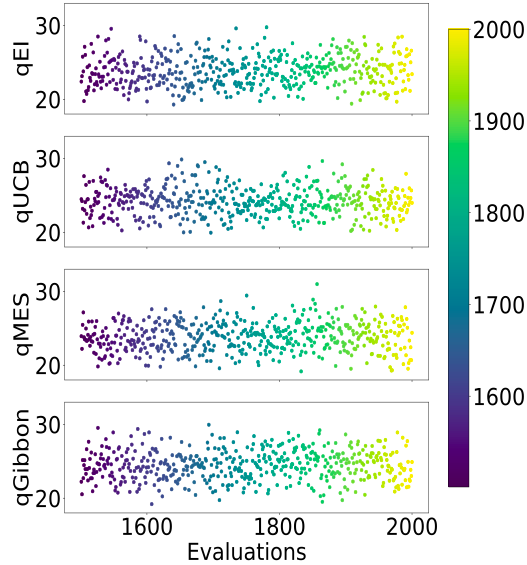}
    \caption{MDPE}
    \label{fig:MDPE_robot}
\end{subfigure}

\caption{Sampling Core—Robot Pushing 14D}
\label{fig:sampling_robot}
\end{figure}

Figure~\ref{fig:sampling_robot} presents the PCE and MDPE from the \textbf{Sampling Core Metrics} for the 14-dimensional Robot Pushing problem. By leveraging PCE plots during local convergence, users can actively guide the optimization process to select points with more diverse coordinate values, thereby reducing the risk of budget inefficiency. For instance, the \textbf{DEEPA} strategy tends to select sample points with a higher range of values in the second dimension compared to the 14th dimension in the final iterations. Additionally, the MDPE plot proves valuable at any stage of the optimization process, providing a clear visualization of the point-wise distances between the evaluated set and the sample points chosen by a given acquisition function. For example, in the last few iterations, \textbf{qMES} tends to focus on points that are closer (with lower distance values) compared to other baselines, making it more exploitative.

\begin{figure}[!]
\centering

\begin{subfigure}{0.35\textwidth}
    \centering
    \includegraphics[width=\linewidth]{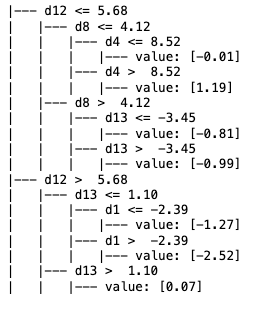}
    \caption{DT rules — 1st Iteration}
\end{subfigure}
\begin{subfigure}{0.35\textwidth}
    \centering
    \includegraphics[width=\linewidth]{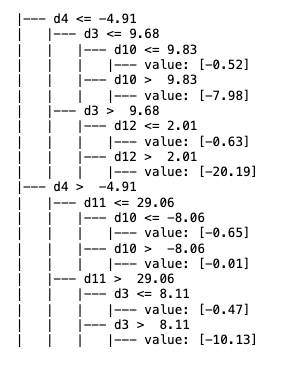}
    \caption{DT rules — 30th Iteration}
\end{subfigure}

\caption{PSSA—Robot Pushing 14D}
\label{fig:pssa_robot}
\end{figure}

\begin{figure}[!]
\centering

\begin{subfigure}{0.5\textwidth}
    \centering
    \includegraphics[width=\linewidth]{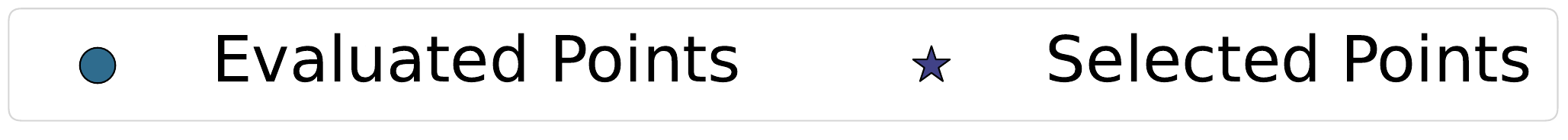}
\end{subfigure}

\vspace{0.5em}

\begin{subfigure}{0.4\textwidth}
    \centering
    \includegraphics[width=\linewidth]{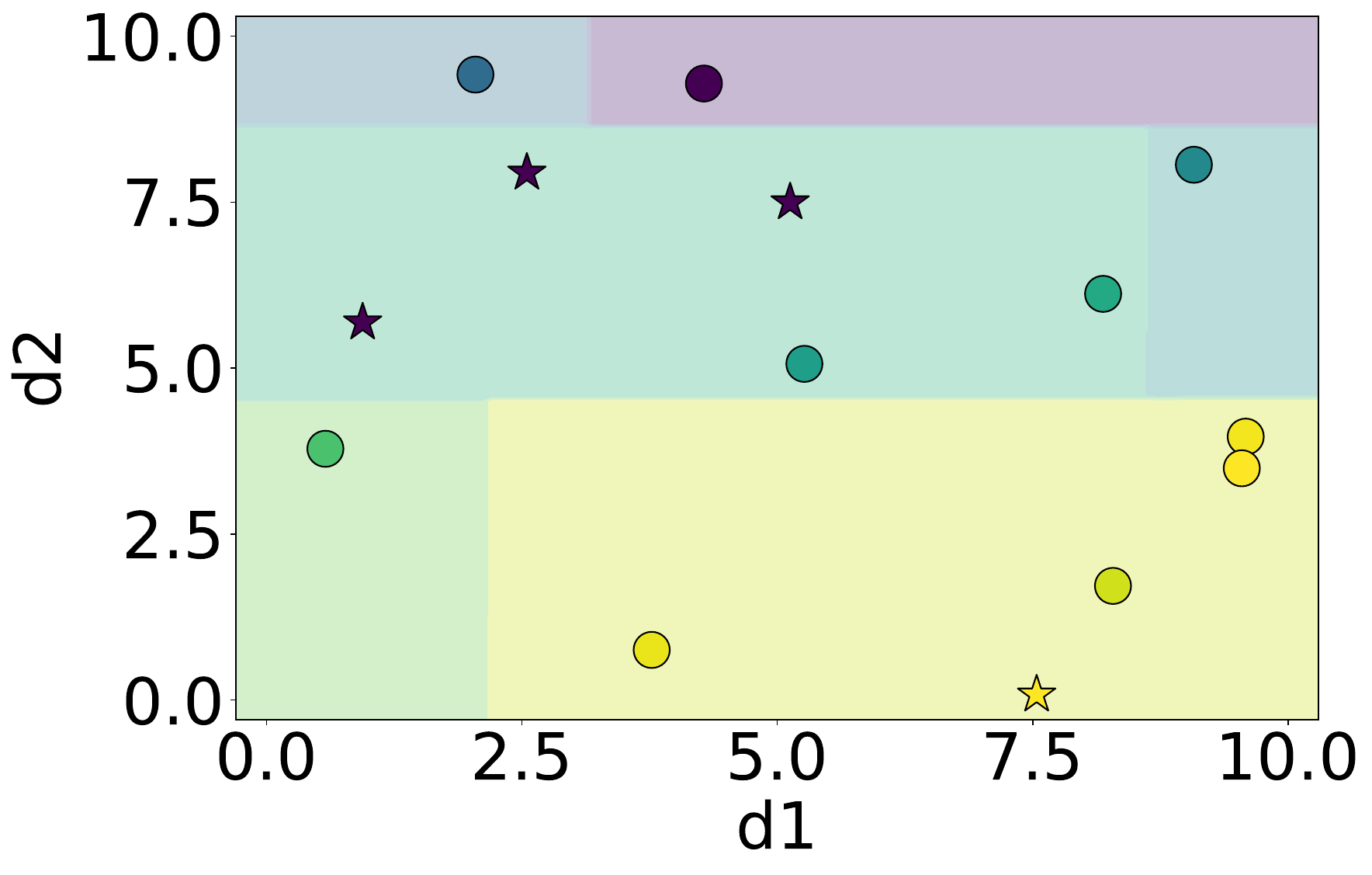}
    \caption{Partitions — 1st iteration}
\end{subfigure}
\begin{subfigure}{0.4\textwidth}
    \centering
    \includegraphics[width=\linewidth]{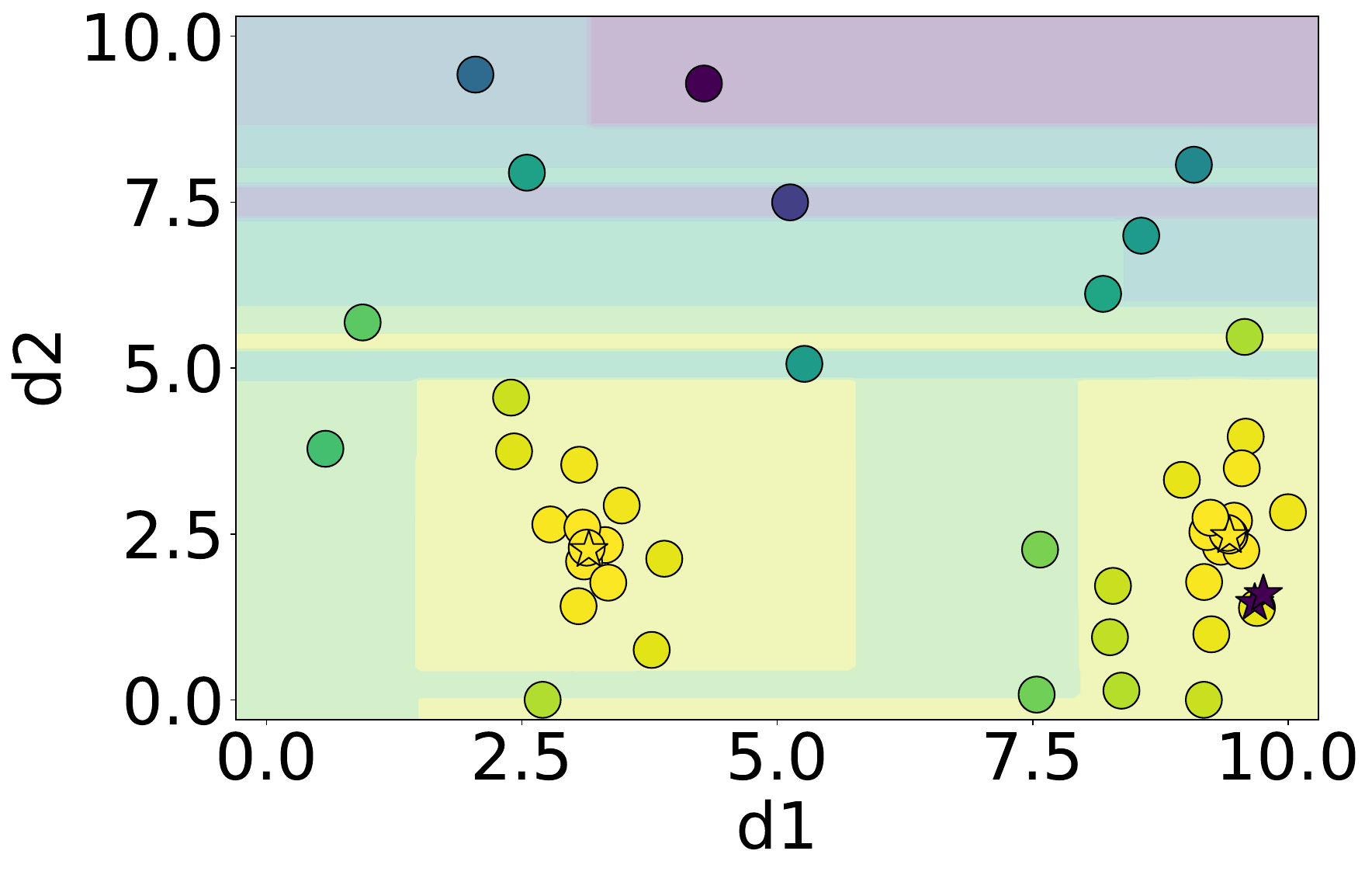}
    \caption{Partitions — 10th iteration}
\end{subfigure}

\caption{PSSA—Branin 2D}
\label{fig:pssa_branin}
\end{figure}

\begin{figure}[!]
\centering

\begin{subfigure}{0.8\textwidth}
    \centering
    \includegraphics[width=\linewidth]{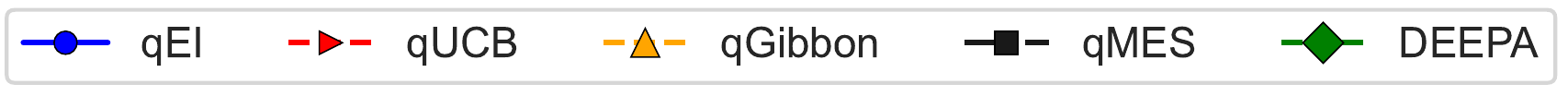}
\end{subfigure}

\vspace{0.5em}

\begin{subfigure}{0.4\textwidth}
    \centering
    \includegraphics[width=\linewidth]{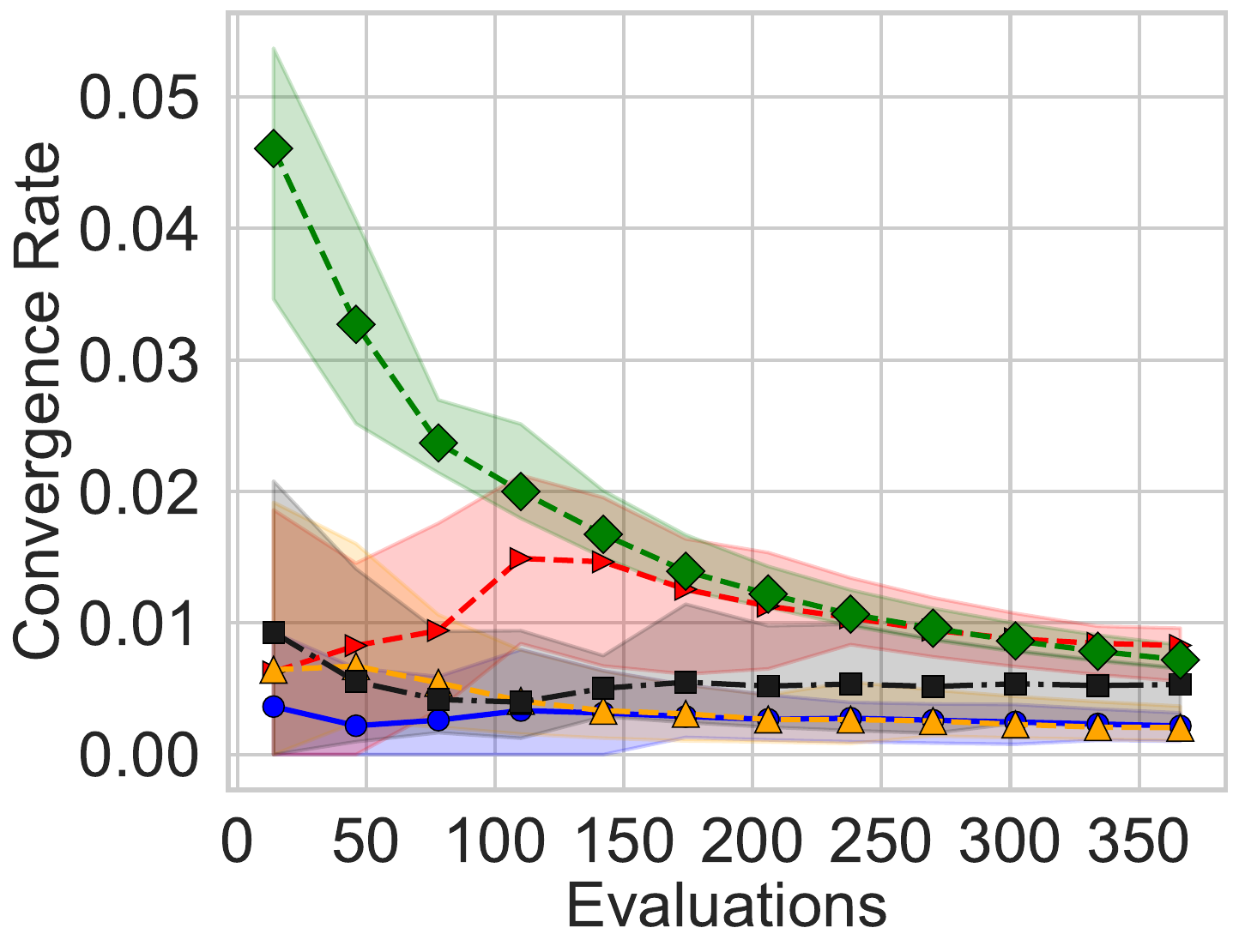}
    \caption{CR}
    \label{fig:CR_levy}
\end{subfigure}
\begin{subfigure}{0.4\textwidth}
    \centering
    \includegraphics[width=\linewidth]{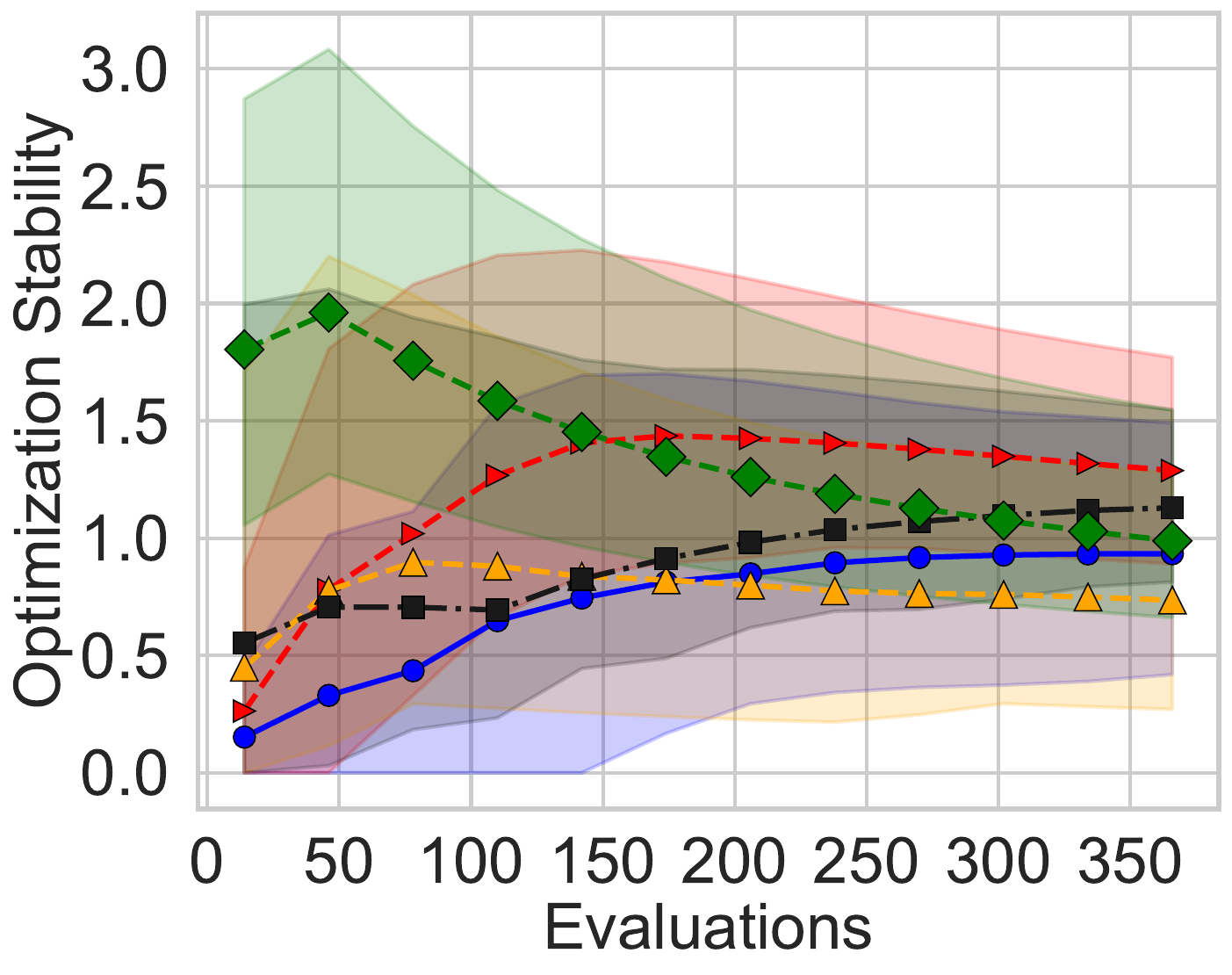}
    \caption{OS}
    \label{fig:OS_levy}
\end{subfigure}

\caption{Optimization Process—Levy 6D}
\label{fig:opt_levy}
\end{figure}

Figures~\ref{fig:pssa_robot} and ~\ref{fig:pssa_branin} depict the PSSA from the \textbf{Optimization Process Metrics} for the Branin and Robot Pushing problems. PSSA enables users to monitor promising regions within the solution space (highlighted in yellow in Figure~\ref{fig:pssa_branin}
) and locate newly selected sample points by revealing the partitioning rules (Figure~\ref{fig:pssa_robot}). For example, in the Robot Pushing problems, we can calculate the number of points selected from each partition based on the Decision Tree (DT) rules in each iteration ($k=50$). Additionally, PSSA assists users in adjusting the sampling strategy if excessive exploitation or exploration is detected within specific areas of the solution space. Moreover, plots in Figure~\ref{fig:opt_levy} provide insight into the performance of different SO baselines. For the 6-dimensional Levy, \textbf{DEEPA} outperforms other baselines in terms of convergence rate, while \textbf{qGibbon} shows greater stability (less variation) across the final iterations.

\begin{figure}[!]
\centering

\begin{subfigure}{0.4\textwidth}
    \centering
    \includegraphics[width=\linewidth]{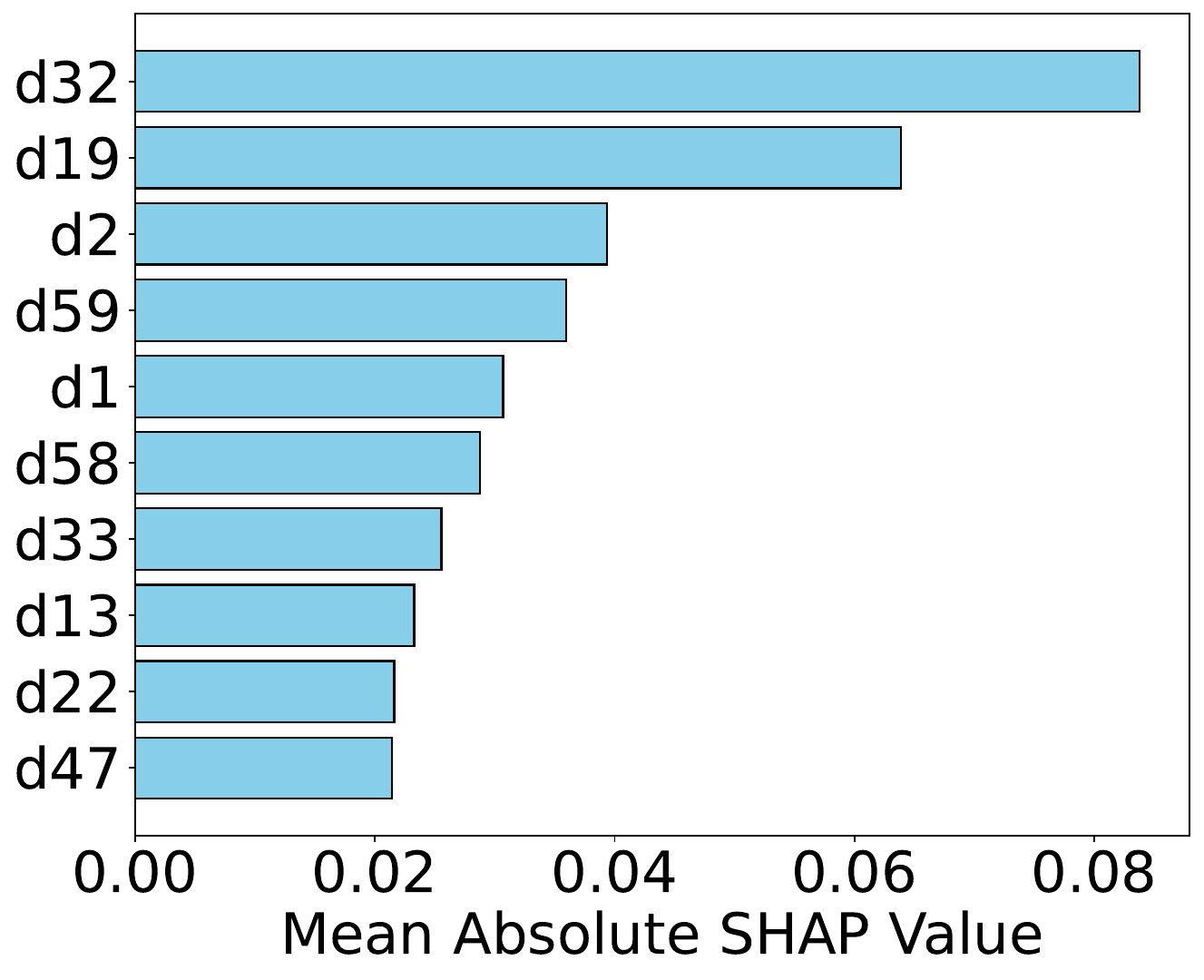}
    \caption{FIEE Top 10 features}
    \label{FIEE-rover}
\end{subfigure}
\begin{subfigure}{0.4\textwidth}
    \centering
    \includegraphics[width=\linewidth]{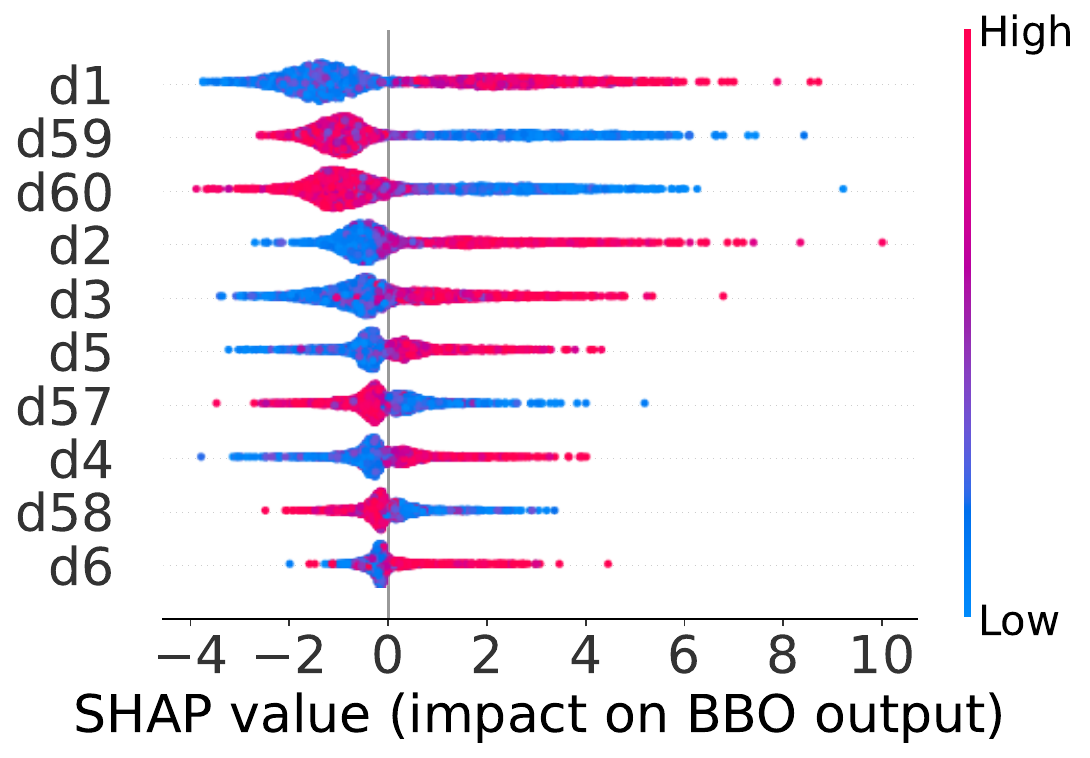}
    \caption{FIBB Top 10 features}
    \label{FIBB-rover}
\end{subfigure}

\caption{Feature Importance—Rover 60D (DEEPA)}
\label{fig:FI_Rover}
\end{figure}

Figures~\ref{FIEE-rover} and ~\ref{FIBB-rover}
show the top 10 features (coordinates) utilizing FIBB and FIEE (exploration) from the \textbf{Feature Importance Metrics} for the high-dimensional Rover Trajectory test problem. For example, dimensions 32 and 19 significantly contribute to the BBO exploration objective (Maxmin distance) while the first and last 3 dimensions have more impact on the true black-box objective value. 

\begin{figure}[!]
\centering

\begin{subfigure}{0.4\textwidth}
    \centering
    \includegraphics[width=\linewidth]{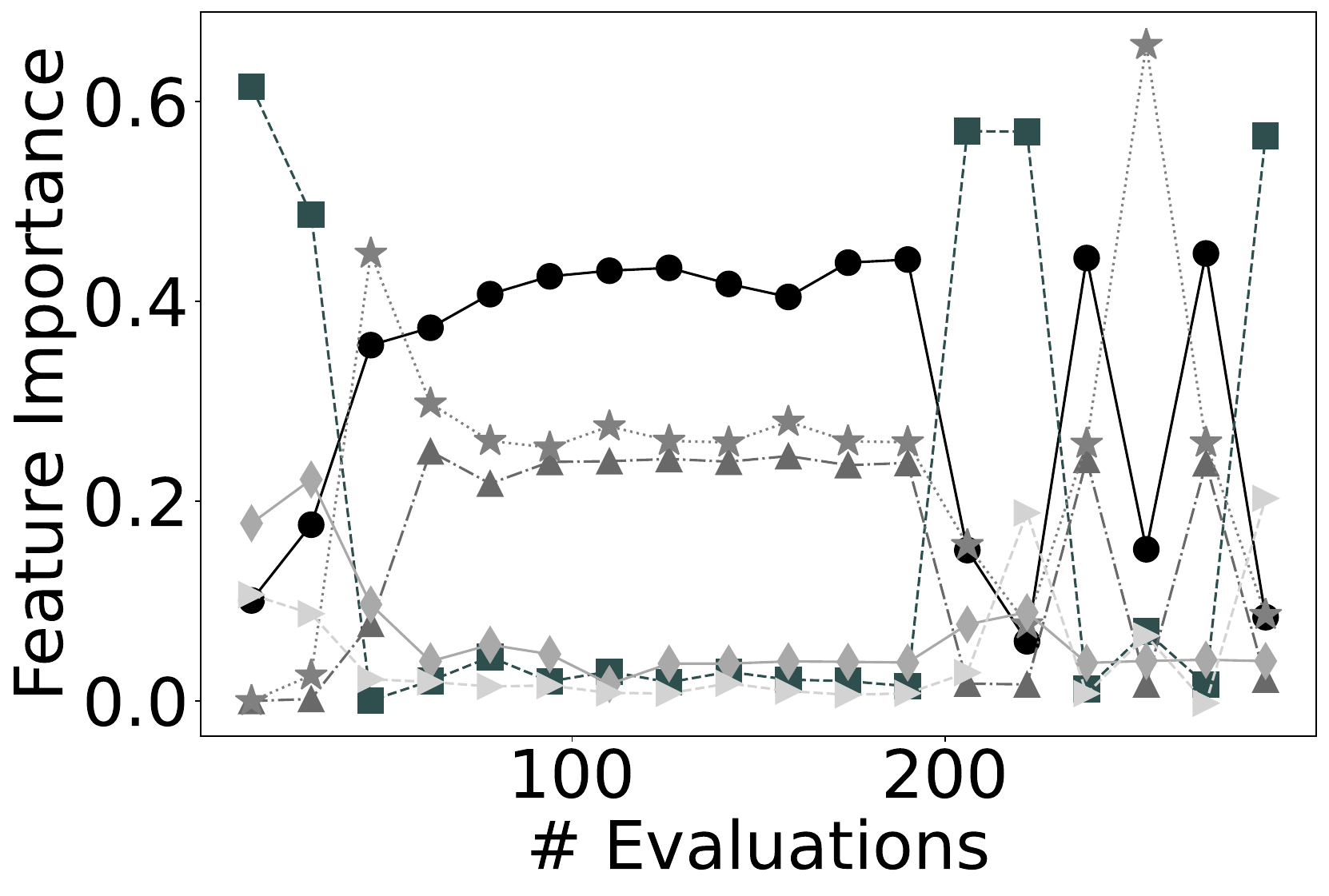}
    \caption{Srg: GP — FI: MARS}
    \label{fig:levyGP}
\end{subfigure}
\begin{subfigure}{0.4\textwidth}
    \centering
    \includegraphics[width=\linewidth]{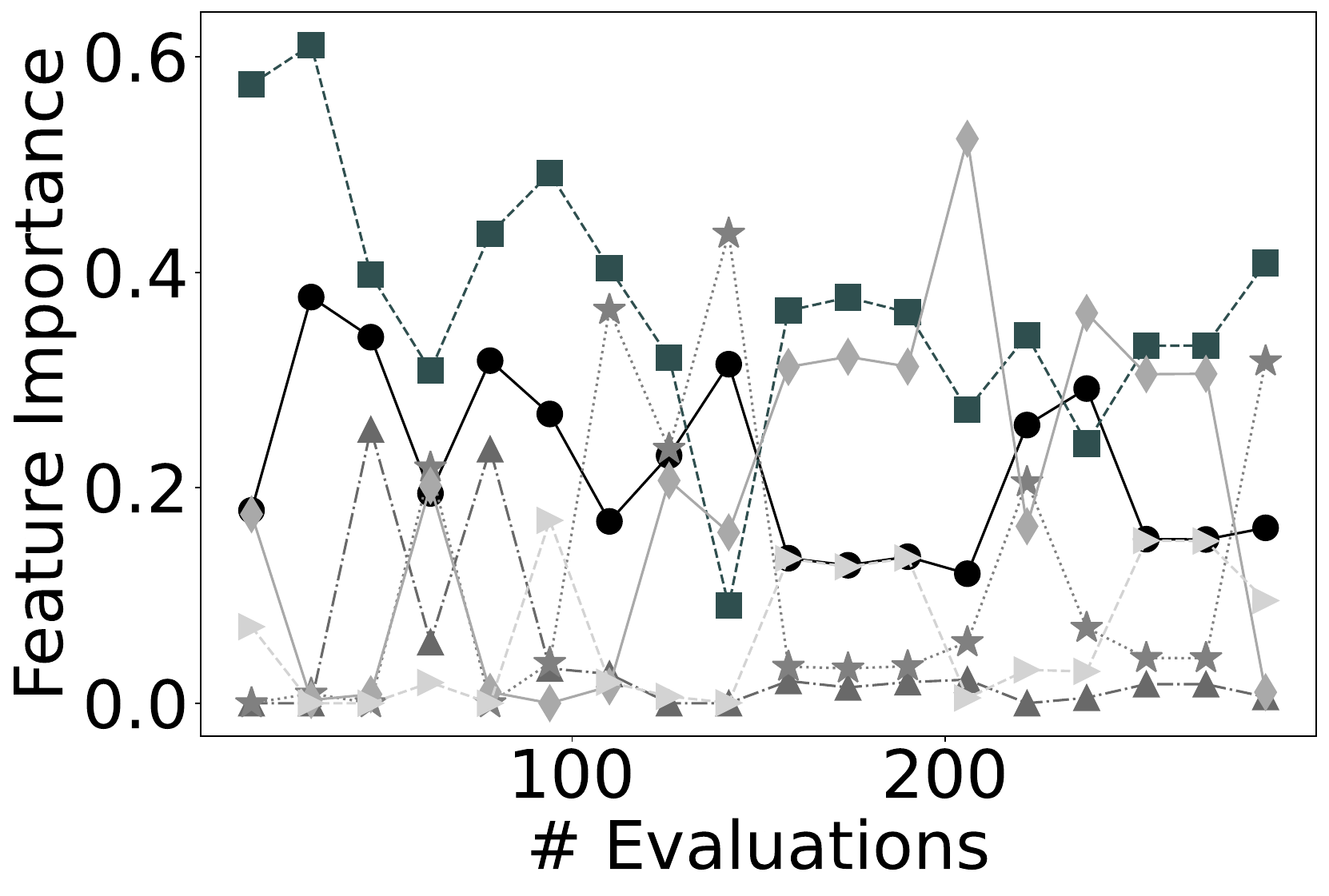}
    \caption{Srg: MARS — FI: MARS}
    \label{fig:levymars}
\end{subfigure}

\caption{FIS—Levy 6D (DEEPA)}
\label{fig:FIS}
\end{figure}

Moreover, Figures~\ref{fig:levyGP} and ~\ref{fig:levymars} show the FIS metric for the 6-dimensional Levy test problem. Figure~\ref{fig:levyGP} utilizes GP as the surrogate model and calculates MARS feature importance based on the observed data in each iteration. Figure~\ref{fig:levymars} uses MARS, which is an explainable surrogate model, and utilizes the surrogate's feature importance in each iteration. The user can utilize these results to monitor how the surrogate's important coordinates change across iterations.

\noindent \textbf{Practical Applications of IEMSO Metrics:}

\begin{figure}[!t]
\centering

\begin{subfigure}{0.24\textwidth}
    \centering
    \includegraphics[width=\linewidth]{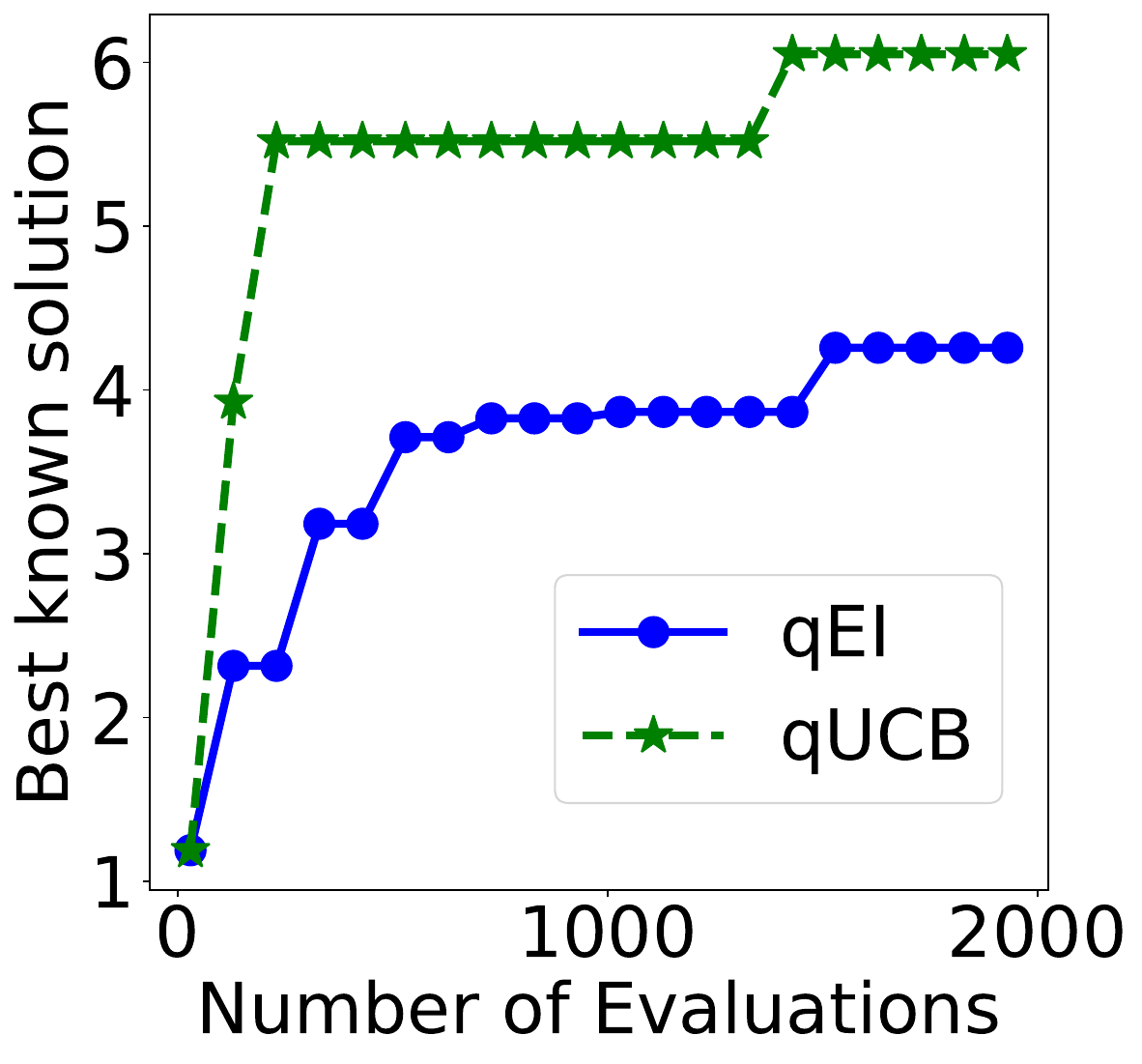}
    \caption{Convergence}
    \label{robot-bks}
\end{subfigure}\hfill
\begin{subfigure}{0.24\textwidth}
    \centering
    \includegraphics[width=\linewidth]{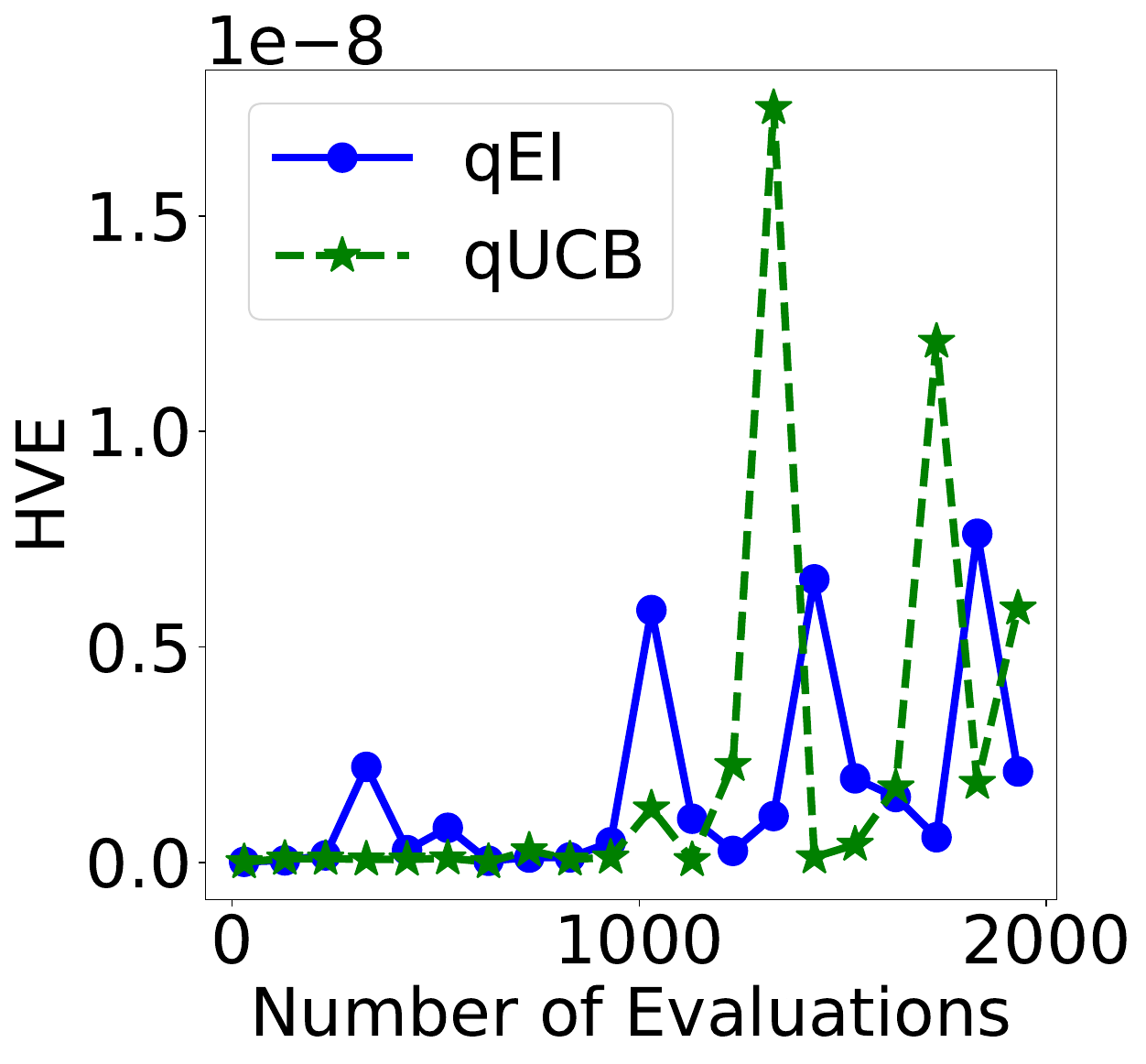}
    \caption{HVE}
    \label{robot-hve}
\end{subfigure}\hfill
\begin{subfigure}{0.24\textwidth}
    \centering
    \includegraphics[width=\linewidth]{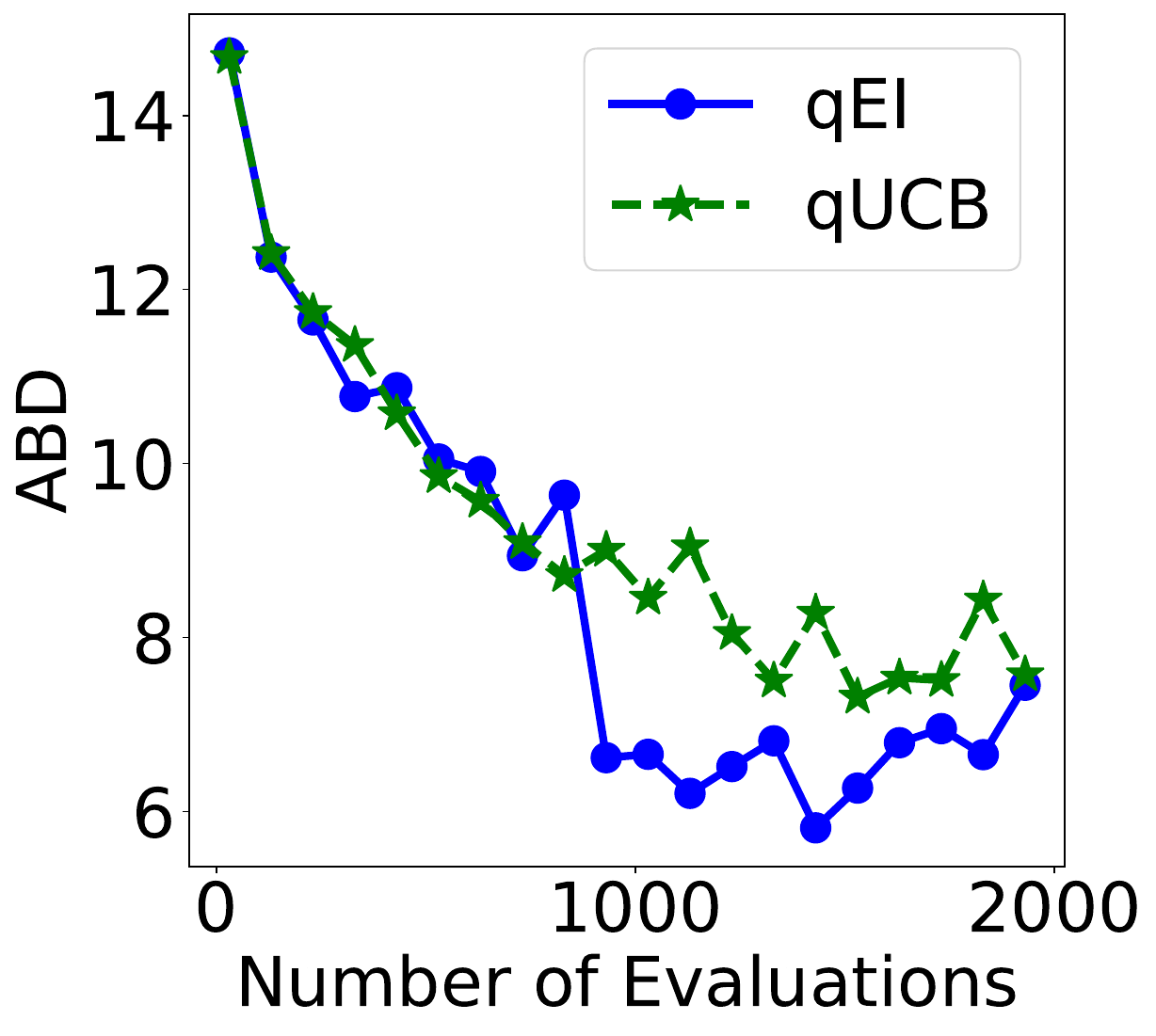}
    \caption{ABD}
    \label{robot-abd}
\end{subfigure}\hfill
\begin{subfigure}{0.24\textwidth}
    \centering
    \includegraphics[width=\linewidth]{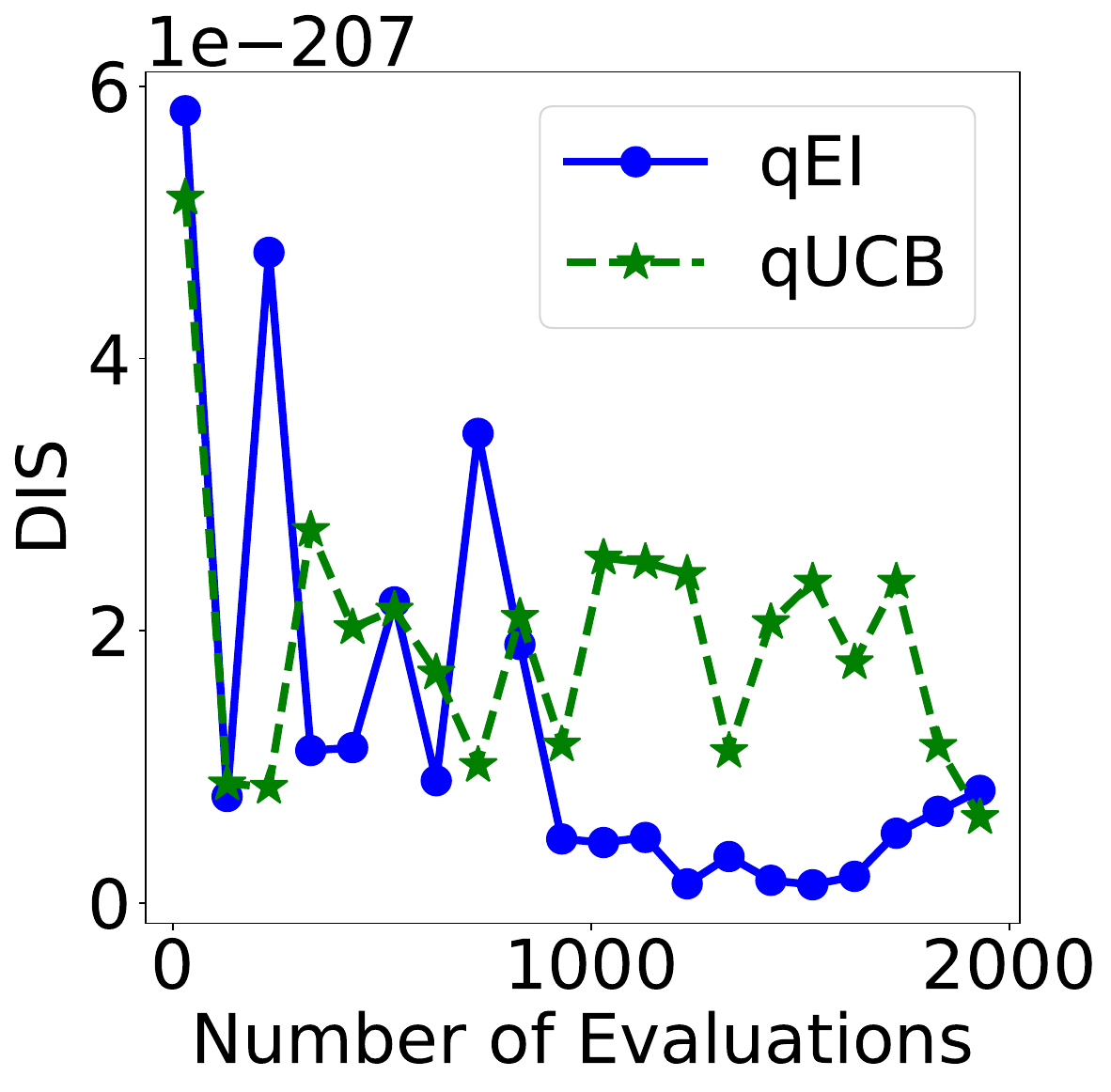}
    \caption{DIS}
    \label{robot-dis}
\end{subfigure}

\caption{Batch Properties Metrics for Performance Comparison: Robot Pushing Problem (14D)}
\label{exp:robot-c}
\end{figure}

\begin{figure}[!t]
\centering

\begin{subfigure}{0.24\textwidth}
    \centering
    \includegraphics[width=\linewidth]{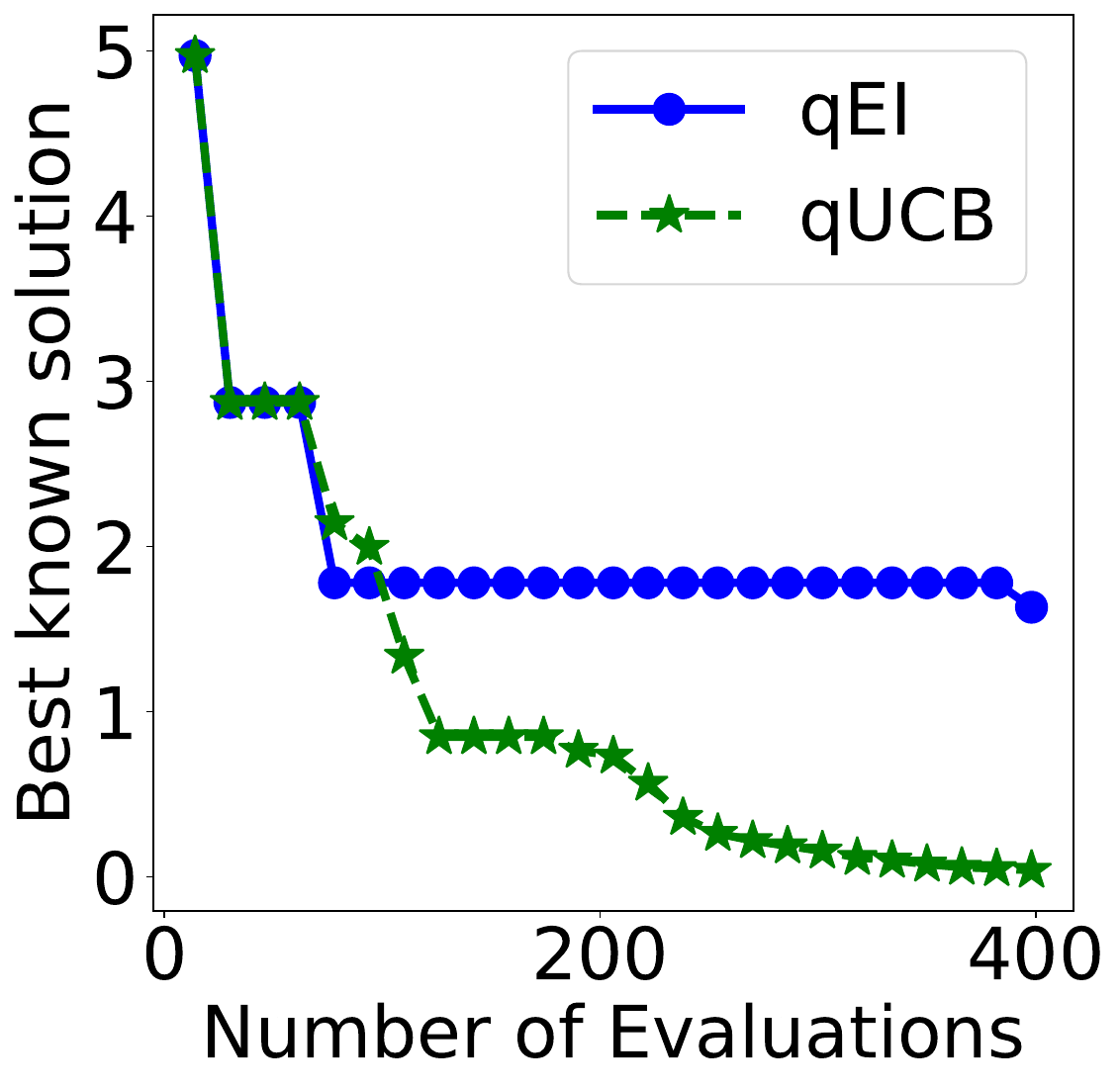}
    \caption{Convergence}
    \label{levy-bks}
\end{subfigure}\hfill
\begin{subfigure}{0.24\textwidth}
    \centering
    \includegraphics[width=\linewidth]{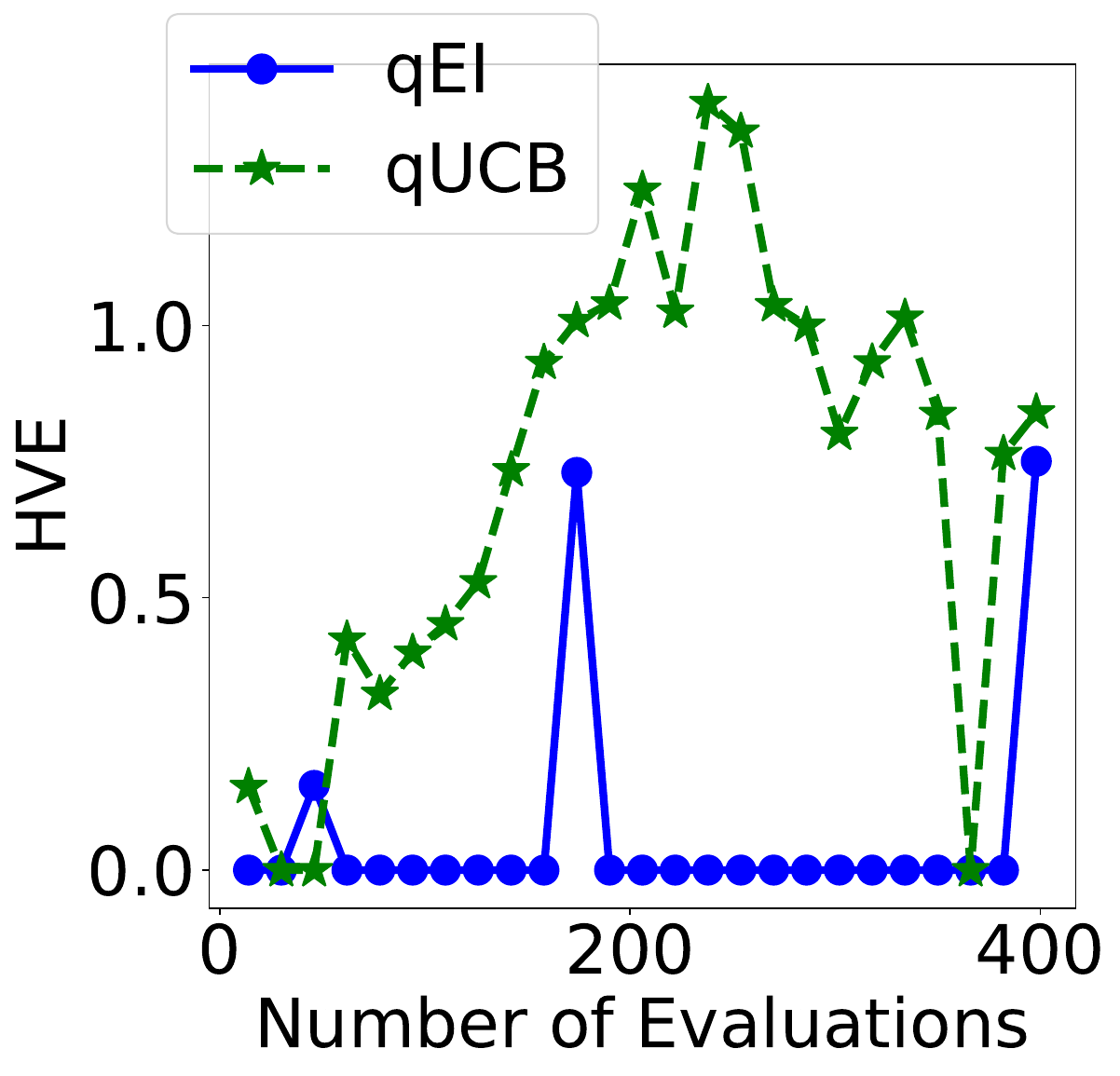}
    \caption{HVE}
    \label{levy-hve}
\end{subfigure}\hfill
\begin{subfigure}{0.24\textwidth}
    \centering
    \includegraphics[width=\linewidth]{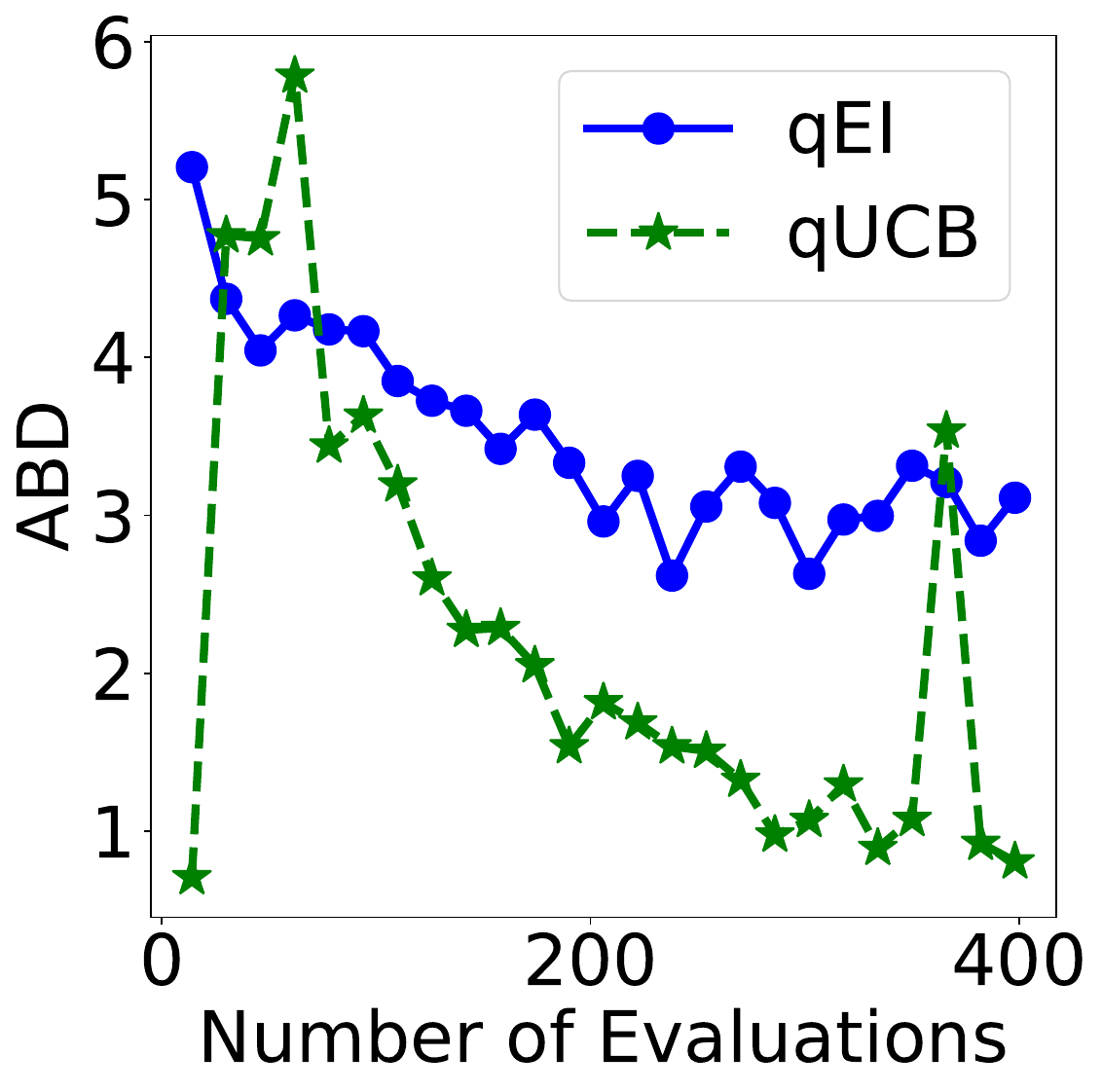}
    \caption{ABD}
    \label{levy-abd}
\end{subfigure}\hfill
\begin{subfigure}{0.24\textwidth}
    \centering
    \includegraphics[width=\linewidth]{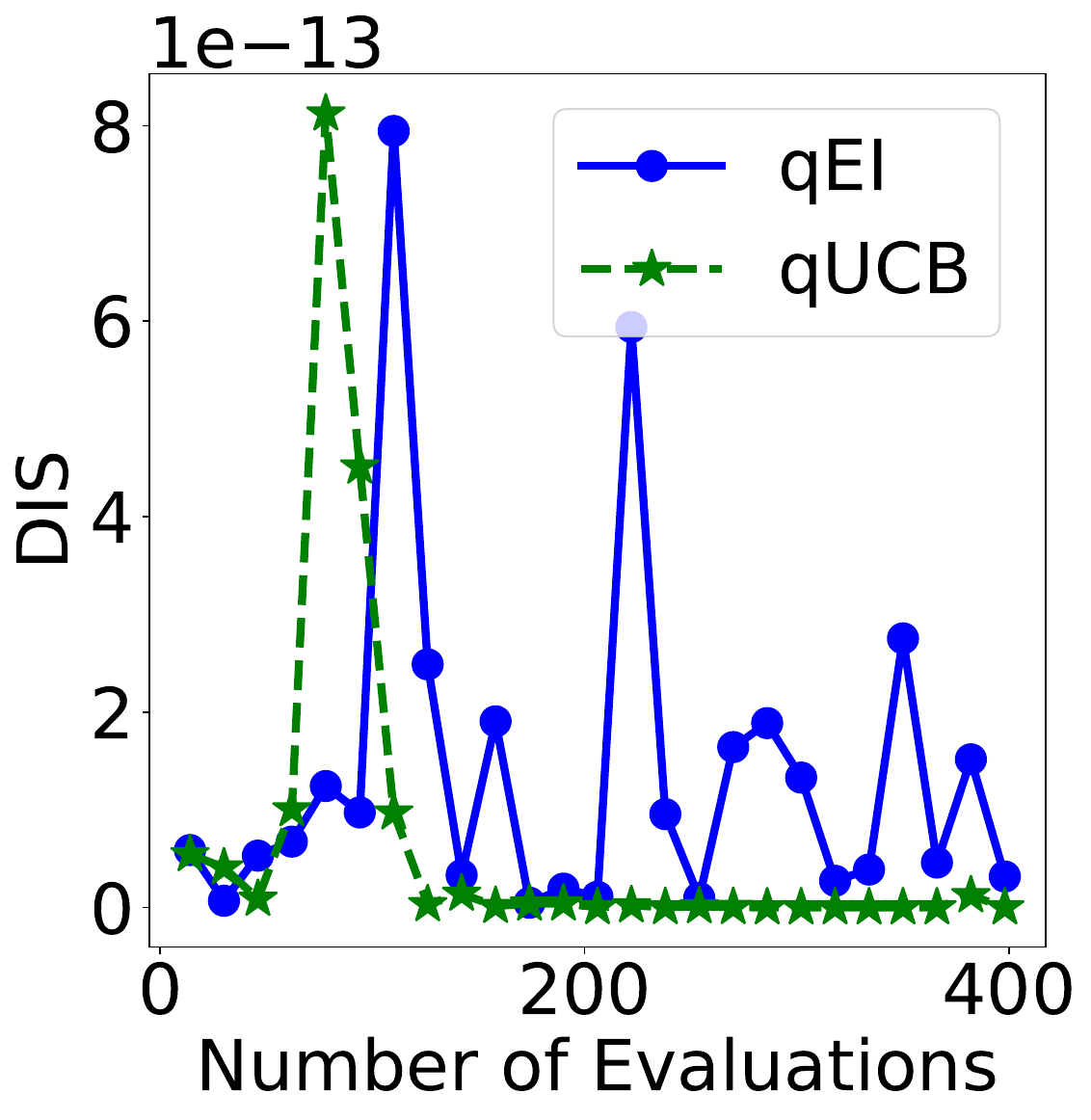}
    \caption{DIS}
    \label{levy-dis}
\end{subfigure}

\caption{Batch Properties Metrics for Performance Comparison: Levy Test Problem (6D)}
\label{exp:levy-c}
\end{figure}

To further illustrate the application of IEMSO metrics, we examine two comparative examples of using \textbf{qEI} and \textbf{qUCB} on the same initialization run of the real-world Robot Pushing problem and Levy test problem. Figures ~\ref{exp:robot-c} and ~\ref{exp:levy-c} present the batch property metrics, including DIS, ABS, and HVE, alongside the typical convergence plot, which reflects the best-observed solution after each iteration.

In standard practice, SO practitioners primarily rely on 
convergence pattern, as shown in Figure~\ref{robot-bks} and Figure~\ref{levy-bks}, which depicts the best-known solution and the corresponding improvement in the black-box objective following expensive evaluations.
By using these performance-based plots—alternatively, regret could be plotted—users can compare the outcomes of \textbf{qEI} and \textbf{qUCB}. In both examples, it is evident that \textbf{qUCB} achieves faster convergence and a superior final solution, making it more suitable for these applications in the Robot Pushing and Levy test problem. However, the reasoning behind this performance difference and the sampling logic driving the gap often remains unclear.

Using batch property metrics derived from IEMSO, such as HVE, ABD, and DIS, as shown in Figures~\ref{robot-hve} to \ref{robot-dis}, users can gain deeper insights into the results obtained in the Robot Pushing problem. For instance, the HVE metric indicates that \textbf{qUCB} better maintains the exploration-exploitation trade-off compared to \textbf{qEI}. Similarly, DIS and ABD metrics reveal that \textbf{qUCB} emphasizes exploration more as iterations progress, while the exploratory behavior of Expected Improvement diminishes over time. Additionally, \textbf{qUCB} exhibits higher exploitation early on, whereas \textbf{qEI} prioritizes exploration due to greater uncertainty values. Consequently, \textbf{qUCB} selects points that are farther from the evaluated set and are more diverse than those chosen by \textbf{qEI}.

In the context of the Robot Pushing problem, the enhanced batch diversity and exploration by \textbf{qUCB} significantly improve the exploration-exploitation balance, leading to its superior performance over \textbf{qEI}. However, this logic does not extend to interpreting the results obtained for the Levy problem. In contrast, the IEMSO batch property metrics such as HVE, ABD, and DIS, as shown in Figures~\ref{levy-hve} to Figure~\ref {levy-dis}, reveal a different pattern. While \textbf{qUCB} maintains a better exploration-exploitation trade-off, \textbf{qEI} demonstrates increased exploration compared to \textbf{qUCB} as the iterations progress. Instead of focusing on promising regions, \textbf{qEI} prioritizes reducing uncertainty and exploring more extensively. This behavior ultimately leads to \textbf{qUCB} outperforming \textbf{qEI} in this scenario as well.

\section{Conclusion}

In this paper, we introduced Inclusive Explainability Metrics for Surrogate Optimization (IEMSO), a comprehensive framework designed to enhance transparency and build trust in black-box optimization (BBO) methods. Our approach addresses significant gaps in the current state of explainability for surrogate optimization, providing a set of model-agnostic metrics that can be applied across different surrogate frameworks and optimization scenarios.
Through extensive experimental analysis on both synthetic and real-world benchmark problems, we demonstrated the effectiveness of IEMSO in offering valuable insights into various stages of the optimization process. The metrics were divided into four main categories: Sampling Core Metrics, Batch Properties Metrics, Optimization Process Metrics, and Feature Importance Metrics. Each category serves a unique purpose in elucidating different aspects of surrogate optimization, from explaining sampling strategies to understanding feature importance and evaluating the overall optimization process.
Our results show that IEMSO not only enhances the explainability of surrogate optimization outcomes but also enables experts to understand the reason behind the performance difference among various approaches. By offering both intermediate and post-hoc explanations, IEMSO ensures that the decision-making processes are transparent, interpretable, and trustworthy, thereby increasing the reliability and applicability of surrogate optimization in complex, real-world scenarios. Future work could focus on expanding the set of explainability metrics and developing a user-friendly interface to further enhance the tool's applicability.

\bibliographystyle{apalike}
\bibliography{ref}

\section*{Biographies}
\fontsize{10}{12}\selectfont

Nazanin Nezami is a researcher at the University of Illinois Chicago. Her research interests include black-box optimization, surrogate modeling, and explainability in machine learning-driven decision systems.

Hadis Anahideh is an Assistant Professor at the University of Illinois Chicago. Her research focuses on black-box optimization, sequential decision-making under uncertainty, responsible machine learning, and explainable AI.

\end{document}